\documentclass{bmvc2k}

\title{FairReL: Deepfake Detection using Fairness-Aware Representation Learning}

\addauthor{Xiaoman Lu}{xiaoman.lu@warwick.ac.uk}{1}
\addauthor{Jiaqi Li}{Jiaqi.Li.16@warwick.ac.uk}{1}
\addauthor{Shuntian Zheng}{Shuntian.Zheng@warwick.ac.uk}{1}
\addauthor{Huiping Chen}{h.chen.13@bham.ac.uk}{2}
\addauthor{Yu Guan\textsuperscript{*}}{Yu.Guan@warwick.ac.uk}{1}

\addinstitution{
 Department of Computer Science\\
 University of Warwick\\
 Coventry, United Kingdom
}
\addinstitution{
 School of Computer Science\\
 University of Birmingham\\
 Birmingham, United Kingdom
}

\runninghead{Lu et al.}{FairReL: Fairness-Aware Representation Learning}

\def\etal{\emph{et al}\bmvaOneDot}

\usepackage{graphicx}
\usepackage{booktabs}
\usepackage{multirow}
\usepackage{amsmath,amssymb}
\usepackage{algorithm}
\usepackage{algpseudocode}
\usepackage{enumitem }
\usepackage{threeparttable}

\usepackage{placeins} 

\usepackage{xcolor}
\usepackage{colortbl}
\definecolor{ourrow}{gray}{0.9}   

\newcommand{\FFPR}{F_{\mathrm{FPR}}}
\newcommand{\FMEO}{F_{\mathrm{MEO}}}

\newcommand{\loss}{\mathcal{L}}

\begin{document}

\maketitle
{\let\thefootnote\relax\footnotetext{\textsuperscript{*}Corresponding author.}}

\begin{abstract}
Although recent deepfake detectors achieve high overall accuracy, their errors remain unevenly distributed across demographic subgroups, with real faces from certain groups more often misclassified as fake. Existing fairness-aware detectors typically regularise the entire feature representation, without identifying or controlling the specific components that drive unfair predictions. Such coarse intervention can over-suppress useful forgery cues while leaving demographic structure in component-specific subspaces. To address this, we identify two subgroup-sensitive components: multi-scale spatial features, which encode local facial and forgery patterns, and fine-tuning-induced residual features, which adapt the backbone to the unfair training distribution. We propose FairReL, a fairness-aware representation-learning framework that targets both components with dedicated demographic supervision. FairReL uses an SVD-decomposed foundation-model backbone to isolate the fine-tuning-induced residual representation, and introduces two complementary losses. Group-Conditional Wavelet Decorrelation (GCWD) suppresses subgroup-imbalanced structure across spatial wavelet sub-bands, while Subspace-Localised Mean Alignment (SLMA) aligns subgroup means within each real/fake class in the residual representation. Experiments on FF++, Celeb-DF, DFD and DFDC show that, against the state-of-the-art fairness-aware detector, FairReL improves unseen-dataset AUC by 3.9\% while reducing subgroup FPR disparity by 10.2\%.
Code is available at \url{https://github.com/xiaoman89/FairReL}.
\end{abstract}

\section{Introduction}
\label{sec:intro}
With rapid advances in generative AI, deepfakes have become increasingly realistic and accessible \cite{pei2026deepfake}. 
These synthetic facial images and videos can be produced by modern generative models, including variational autoencoders~\cite{kingma2013auto,sohn2015learning}, generative adversarial networks~\cite{goodfellow2014generative,karras2019style,karras2020analyzing}, and diffusion models~\cite{dhariwal2021diffusion}, raising serious societal concerns~\cite{pei2026deepfake}. 
In response, extensive research has focused on deepfake detection, with recent detectors achieving strong overall accuracy on standard benchmarks~\cite{chollet2017xception,li2020face,qian2020thinking} and showing improved generalisation to manipulation methods unseen during training~\cite{yan2023ucf,yan2024effort,yermakov2026deepfake}.

\begin{figure}[t]
    \centering
    \includegraphics[width=0.76\columnwidth]{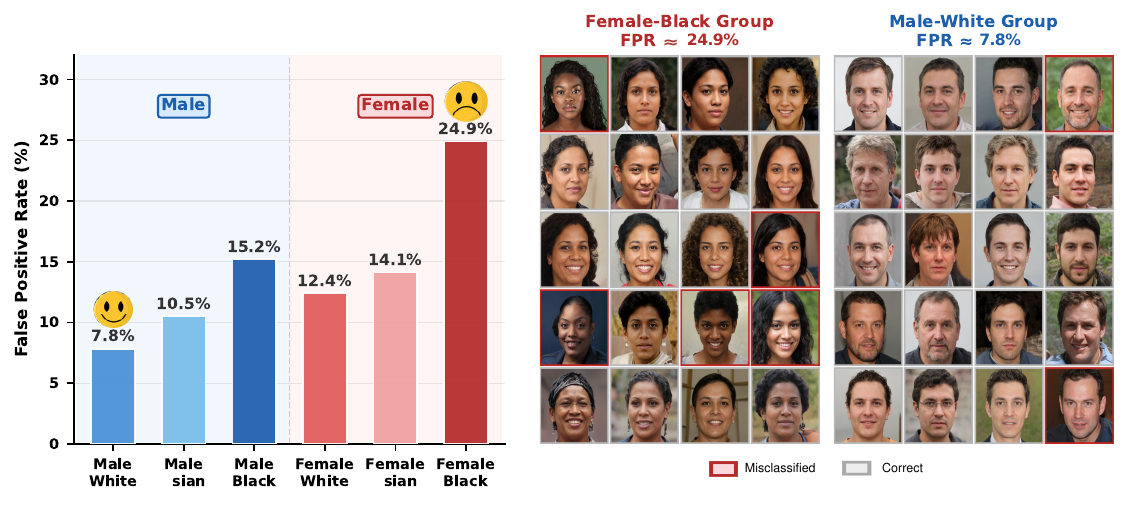}
    \caption{%
   False positive rate (the fraction of real faces misclassified as fake, the lower the better) across six demographic subgroups for a vanilla Xception detector~\cite{chollet2017xception}, evaluated at a single shared threshold. The Female-Black group shows the highest rate.
    }
    \label{fig:demographic_gap}
\end{figure}

However, recent studies have shown that current deepfake detectors exhibit demographic unfairness~\cite{trinh2021examination,nadimpalli2022gbdf,masood2023deepfakes}, with detection errors unevenly distributed across groups such as gender and ethnicity~\cite{xu2024analyzing}.
As shown in Figure~\ref{fig:demographic_gap}, a vanilla Xception detector~\cite{chollet2017xception} misclassifies real Female-Black faces as fake over three times as often as real Male-White faces.
This calls for a fairness-aware deepfake detector that reduces subgroup-level error gaps without sacrificing detection accuracy, while preserving such fairness across manipulation methods and demographic distributions unseen during training.

Existing fairness-aware deepfake detectors mainly reduce subgroup disparity by adding fairness constraints during training. 
Some methods operate on the final prediction by reweighting samples or penalising group-specific error gaps~\cite{ju2024improving,lin2024preserving,cheng2026fair}, while others regularise learned features by matching their distributions across demographic groups \cite{ravfogel2020null,ravfogel2022linear,madras2018learning,edwards2016censoring}. 
However, both strategies intervene only after multiple sources of evidence have already been mixed together, and therefore provide limited control over what is actually being suppressed.
In prediction-level methods, subgroup-sensitive shortcuts may remain encoded in the representation even if the final outputs appear more balanced, making the fairness gain brittle under distribution shift \cite{lin2024preserving,cheng2026fair}. 
In representation-level methods, applying the same regularisation to the entire feature can blur the distinction between harmful subgroup cues and useful forgery evidence, leading to a trade-off between fairness and detection performance \cite{edwards2016censoring}. 
This suggests that the key question is not simply whether to impose a fairness constraint, but where subgroup-sensitive structure arises and how it can be targeted more precisely.

Building on this gap, our motivation is to control two sources of unfairness in deepfake detectors. 
The first source comes from the spatial-scale mismatch between demographic appearance and forgery evidence.
Many subgroup-related facial attributes (e.g., skin tone, overall facial geometry) are expressed as relatively coarse and spatially smooth patterns, whereas many forgery cues arise from more local inconsistencies (e.g., blending boundaries, texture artefacts, and synthesis traces) ~\cite{qian2020thinking}. 
This scale difference suggests that decomposing the spatial feature map into frequency bands can provide a useful handle for separating part of the demographic-related appearance variation from forgery-related evidence. 
The second source arises during task adaptation. 
Modern deepfake detectors are commonly fine-tuned on benchmarks whose demographic coverage is limited or unevenly distributed \cite{xu2024analyzing}. 
When a pre-trained backbone is adapted on such data, subgroup-specific correlations can be absorbed into the task-adapted representation and later be mistaken for evidence of fakeness \cite{ju2024improving,trinh2021examination}.

Motivated by these two bias sources, we propose \textbf{FairReL}, a fairness-aware representation-learning framework for generalisable deepfake detection.
Instead of imposing a single fairness constraint on the final embedding, FairReL intervenes at the stages where subgroup bias is most likely to enter and be amplified. 
First, before global pooling collapses spatial structure, \textbf{Group-Conditional Wavelet Decorrelation (GCWD)} decomposes the spatial feature map into wavelet sub-bands and penalises subgroup-dependent spectral imbalance. 
This alleviates demographic appearance shortcuts at the spatial-frequency level while preserving local forgery evidence. 
Second, to prevent biased fine-tuning from contaminating the whole backbone, FairReL decomposes each adapted weight into a frozen principal component and a trainable residual component. 
\textbf{Subspace-Localised Mean Alignment (SLMA)} then aligns subgroup means only in the residual representation and only within the same real/fake class, reducing adaptation-induced demographic shifts without weakening the discriminative space used for detection. 
In this way, FairReL targets two complementary mechanisms of unfairness (spatial-frequency bias and residual adaptation bias) while requiring demographic labels only during training and adding no inference-time overhead.
Extensive experiments demonstrate that FairReL improves the average AUC on unseen datasets by 3.9 percentage points over existing fairness-aware detector, while reducing subgroup FPR disparity by 10.2\% compared to existing fairness-aware detectors.

The main contributions of this paper are as follows.
\begin{itemize}[itemsep=0pt,topsep=0pt,parsep=0pt,leftmargin=10pt]
\item We identify two representation components that are closely linked to unfairness in deepfake detection. One is the spatial feature map, where frequency decomposition helps separate demographic appearance cues from forgery evidence. The other is the fine-tuning residual, where imbalanced training data can introduce subgroup-specific shifts.

\item We propose \textbf{FairReL}, a targeted fairness-aware framework that controls these components without regularising the whole embedding. \textbf{GCWD} reduces subgroup imbalance in the wavelet bands of spatial features, while \textbf{SLMA} aligns class-conditional subgroup means in the residual subspace to suppress adaptation-induced bias.

\item We conduct extensive experiments on FF++~\cite{rossler2019faceforensics}, Celeb-DF~\cite{li2020celebdf}, DFD~\cite{dufour2019dfd} and DFDC~\cite{dolhansky2020dfdc}, showing that FairReL achieves a stronger fairness--generalisation trade-off than ERM, generalisation-oriented and fairness-oriented baselines. Ablations and representation-level analyses further confirm that GCWD and SLMA target complementary sources of subgroup bias and that localising fairness supervision to the trainable residual is effective.

\end{itemize}

\section{Related Work}
\label{sec:related}

\subsection{Generalisable Deepfake Detection}
\label{sec:rw_generalizable}
Early deepfake detectors exploit visual or temporal artefacts, such as warping traces, head-pose inconsistencies, blending boundaries and temporal cues~\cite{li2018warping,yang2019headpose,guera2018temporal,rossler2019faceforensics,li2020face}; while accurate on their training distribution, they tend to degrade on unseen datasets and forgery methods.
More recent work targets generalisable detection through common-forgery features, foundation-model-based detectors, large benchmarks and forgery-semantic decoupling~\cite{yan2023ucf,ojha2023universal,yan2023deepfakebench,yan2024effort,ye2024dfs}.
FairReL sits in this setting but treats fairness as a co-equal requirement: a detector should transfer across forgery methods and demographic groups.

\subsection{Fairness in Deepfake Detection}
\label{sec:rw_fairness}
Fairness in face-related vision systems is an established concern because performance varies across demographic groups~\cite{buolamwini2018gender,grother2019frvt}.
Within deepfake detection, early work examines demographic bias, evaluation protocols and balanced data~\cite{trinh2021examination,pu2022fairness,nadimpalli2022gbdf}, while more recent work introduces algorithm-level fairness objectives and fairness generalisation under domain shift~\cite{ju2024improving,xu2024analyzing,agarwal2024deepfake,lin2024preserving,ezeakunne2024data,cheng2026fair}.
Ju~\etal~\cite{ju2024improving} apply demographic-aware fairness losses at the prediction layer; Lin~\etal~\cite{lin2024preserving} show that source-domain fairness often fails to transfer to unseen datasets; Cheng~\etal~\cite{cheng2026fair} use same-label feature alignment to improve cross-domain robustness.
These methods supervise fairness either at the prediction layer or on the pooled embedding as a whole, treating the representation as a single undifferentiated object.
FairReL departs from this view: it intervenes at two distinct representation levels---multi-scale spectral structure of the spatial feature map, and the trainable residual of the pooled feature---which lets it preserve the transferable visual prior in the principal subspace while still removing adaptation-induced demographic shifts.

\subsection{Fairness-Aware Representation Learning}
\label{sec:rw_adaptation}
Fair representation learning aims to remove subgroup information from learned features via adversarial removal~\cite{edwards2016censoring,madras2018learning}, null-space projection~\cite{ravfogel2020null,ravfogel2022linear} or moment matching~\cite{hardt2016equality}.
These methods usually act on the final embedding, which is coarse for deepfake detection because demographic appearance and forgery evidence can be mixed after pooling. 
Recent studies show that deepfake datasets are demographically uneven, and that detectors fine-tuned on them often inherit subgroup disparities~\cite{trinh2021examination,xu2024analyzing,ju2024improving}. 
This suggests that fairness should also examine the part of the model that changes during fine-tuning. 
Effort~\cite{yan2024effort} provides a useful motivation for this separation by decomposing the backbone with SVD into a frozen principal subspace and a trainable residual subspace. 
However, Effort is designed for generalisation rather than fairness, and therefore does not explicitly constrain unfair subgroup shifts in the residual pathway or separate demographic appearance from forgery evidence in spatial features. 
FairReL addresses these limitations by aligning subgroup means in the class-conditioned residual representation, and by reducing subgroup imbalance in wavelet bands before spatial information is pooled away.

\section{FairReL}
\label{sec:method}

\subsection{Problem Definition}
\label{sec:setup}

Given a deepfake detection dataset $\mathcal{D}=\{(x_i,y_i,\mathbf{s}_i)\}_{i=1}^{N}$ with $N$ samples, each image $x_i$ is associated with a real/fake label $y_i\in\{0,1\}$, where $0$ denotes real and $1$ denotes fake, and a demographic subgroup label $\mathbf{s}_i$. 
Here, $\mathbf{s}_i\in\{0,1\}^{G}$ is a one-hot vector, $G$ denotes the total number of demographic subgroups, and $\sum_{g=1}^{G}s_{i,g}=1$.
At test time, the model receives only the image $x$ and predicts whether it is real or fake.

\textbf{Detector.}
Let $B$ denote the backbone, $f_p$ the pooling operation, and $h_{\phi}$ the binary classifier. 
For an input image $x_i$, the standard detector follows
\begin{equation}
\label{eq:forward}
    \mathbf{F}_i = B(x_i) \in\mathbb{R}^{h\times w\times d},
    \qquad
    \mathbf{z}_i = f_p(\mathbf{F}_i) \in\mathbb{R}^{d},
    \qquad
    \hat{p}_i = h_{\phi}(\mathbf{z}_i) \in [0,1],
\end{equation}
where $\mathbf{F}_i$ is the spatial feature map with height $h$, width $w$, and channel dimension $d$; $\mathbf{z}_i$ is the pooled representation; and $\hat{p}_i$ is the predicted probability that $x_i$ is fake.
A binary prediction is obtained by thresholding $\hat{p}_i$ at an operating point $\tau$, where $\hat{y}_i=1$ if $\hat{p}_i>\tau$ and $\hat{y}_i=0$ otherwise.
This forward path determines where fairness should be imposed. 
Before pooling, $\mathbf{F}_i$ still contains spatial and scale information, which is useful for separating demographic appearance patterns from local forgery evidence. 
After pooling, $\mathbf{z}_i$ becomes compact and prediction-ready, but spatial information has already been mixed.

\begin{figure*}[t]
\begin{center}
\includegraphics[width=1\textwidth]{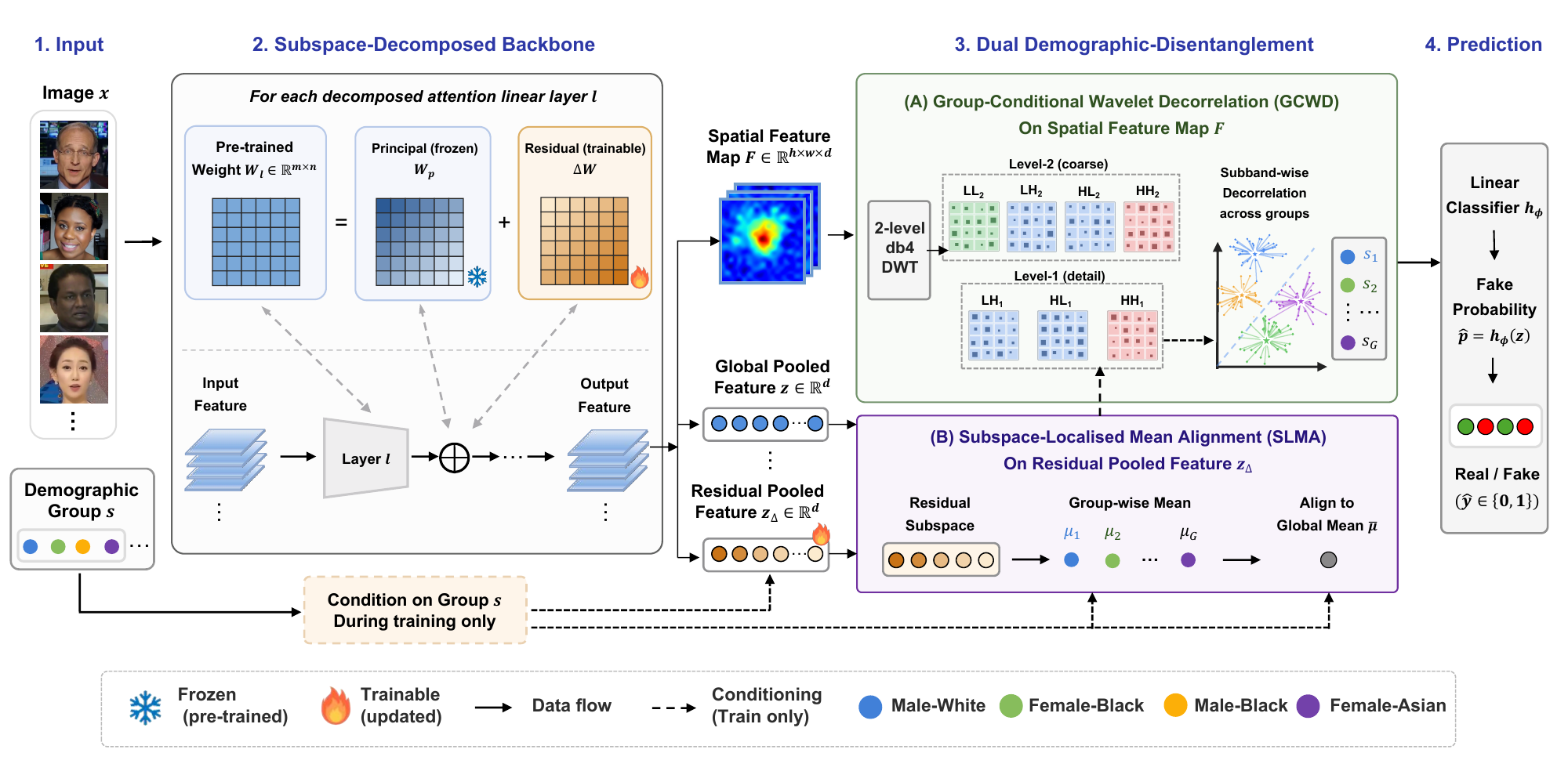}
\end{center}
\caption{%
Training pipeline of FairReL. 
The backbone is decomposed into a frozen principal part $\mathbf{W}_p$ and a trainable residual part $\Delta\mathbf{W}$. 
Given an input image, the standard path produces a spatial feature map $\mathbf{F}$ and a pooled prediction feature $\mathbf{z}$, while the residual-only path produces $\mathbf{z}_{\Delta}$. 
\textbf{(A) GCWD} applies a wavelet transform to $\mathbf{F}$ and reduces subgroup differences in each frequency band. 
\textbf{(B) SLMA} aligns subgroup means in $\mathbf{z}_{\Delta}$ within each real/fake class, so that the residual update is less subgroup-dependent. 
Only $\mathbf{z}$ is fed to the classifier for real/fake prediction. 
At inference, only the backbone and the classifier are retained.
}
\label{fig:method_overview}
\end{figure*}

\textbf{Overview.}
FairReL reframes fairness-aware deepfake detection as a problem of targeted representation control.
We first decompose the backbone $B$ into a frozen principal part and a trainable residual part. 
Instead of regularising the pooled embedding $\mathbf{z}_i$, FairReL applies GCWD to the pre-pooling feature map $\mathbf{F}_i$ to reduce subgroup imbalance in spatial-frequency statistics, and applies SLMA to the fine-tuning residual representation of backbone $B$ to suppress subgroup shifts introduced by imbalanced fine-tuning.
Figure~\ref{fig:method_overview} shows our pipeline.

\subsection{Subspace-Decomposed Backbone}
\label{sec:backbone}
Fine-tuning on demographically imbalanced data can introduce subgroup-specific shifts into the adapted representation. 
A fairness constraint on the full pooled feature $\mathbf{z}_i$ is too coarse, because $\mathbf{z}_i$ contains both the stable visual prior inherited from pre-training and the task-specific changes introduced by deepfake fine-tuning. 
To isolate the latter, FairReL decomposes the backbone into a frozen principal component and a trainable residual component.

Following the SVD decomposition in Effort~\cite{yan2024effort}, each pre-trained linear weight $\mathbf{W}\in\mathbb{R}^{m\times n}$ with $s=\min(m,n)$ singular directions is split as
\begin{equation}
\label{eq:svd_split}
    \mathbf{W} \;=\; \underbrace{\mathbf{U}_{:s-r}\,\boldsymbol{\Sigma}_{:s-r}\,\mathbf{V}_{:s-r}^{\top}}_{\mathbf{W}_{p}\;\text{(frozen, top-$(s-r)$)}}
    \;+\; \underbrace{\mathbf{U}_{s-r:}\,\boldsymbol{\Sigma}_{s-r:}\,\mathbf{V}_{s-r:}^{\top}}_{\Delta\mathbf{W}\;\text{(trainable, rank $r$)}},
\end{equation}
where $\mathbf{U}$ and $\mathbf{V}$ are the left and right singular-vector matrices, and $\boldsymbol{\Sigma}$ is the diagonal singular-value matrix.
Here $:s{-}r$ denotes the top $s{-}r$ singular directions and $s{-}r:$ the remaining $r$, so $\mathbf{W}_{p}$ retains the dominant pre-trained subspace and stays frozen while $\Delta\mathbf{W}$ has rank $r$.
As in~\cite{yan2024effort}, we set $r{=}1$; for ViT-L/14 attention projections this gives frozen rank $1023$ and trainable rank $1$.
We denote the standard backbone using $\mathbf{W}_{p}+\Delta\mathbf{W}$ as $B_{p+\Delta}$, and the residual-only backbone using $\Delta\mathbf{W}$ as $B_{\Delta}$.
Only $\Delta\mathbf{W}$ and the classifier parameters $h_{\phi}$ are trainable.

To keep this split meaningful during training, we follow~\cite{yan2024effort} and penalise overlap between the principal and residual singular-vector bases. 
Let $\hat{\mathbf{U}}=[\mathbf{U}_{:r},\mathbf{U}_{r:}]$ and $\hat{\mathbf{V}}=[\mathbf{V}_{:r},\mathbf{V}_{r:}]$. 
The orthogonality regulariser is
\begin{equation}
\label{eq:orth}
    \loss_{\mathrm{orth}}
    \;=\;
    \bigl\|\hat{\mathbf{U}}^{\top}\hat{\mathbf{U}}-\mathbf{I}\bigr\|_F^{2}
    \;+\;
    \bigl\|\hat{\mathbf{V}}^{\top}\hat{\mathbf{V}}-\mathbf{I}\bigr\|_F^{2},
\end{equation}
where $\mathbf{I}$ is the identity matrix of appropriate dimension.
This term is not a fairness loss.
It preserves the principal/residual separation so that the effect of fine-tuning could be isolated.

\textbf{Two forward passes.}
During training, the standard path uses the full backbone $B_{p+\Delta}$ and produces the features $\mathbf{F}_i$ used for detection.
Since a deep backbone contains attention, normalisation, nonlinearities and layer interactions, the residual representation cannot be reliably obtained by subtracting a principal output from the standard forward pass. 
In addition to \cite{yan2024effort}, to measure the representation change introduced only by fine-tuning, we further run a second path with the frozen principal component $\mathbf{W}_p$ removed (replaced by $\mathbf{0}$):
\begin{equation}
\label{eq:zdelta}
    \mathbf{z}_{\Delta,i}=f_p\bigl(B_{\Delta}(x_i)\bigr)|_{\mathbf{W}_p\,\leftarrow\,\mathbf{0}}.
\end{equation}
The separate residual-only path provides a clean training-time estimate of the adaptation-induced representation, while inference keeps only the standard path.

\subsection{Group-Conditional Wavelet Decorrelation (GCWD)}
\label{sec:gcwd}

\begin{figure}[t]
\centering
\includegraphics[width=\columnwidth]{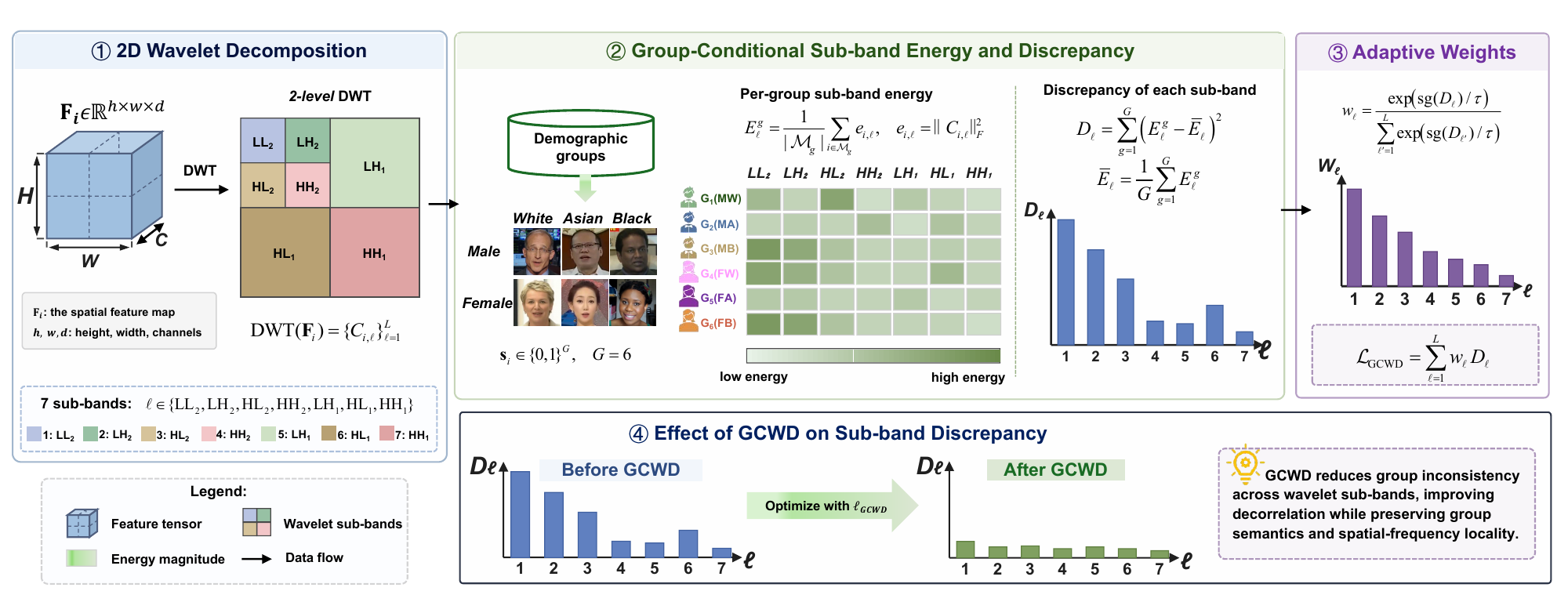}
\caption{%
GCWD computation pipeline. 
\textbf{(1)} The spatial feature map $\mathbf{F}$ is decomposed by DWT into frequency sub-bands, each describing a different scale and direction of spatial variation. 
\textbf{(2)} GCWD computes the average energy of each sub-band within every demographic subgroup and measures the cross-subgroup discrepancy $D_{\ell}$. 
\textbf{(3)} A detached softmax converts $\{D_{\ell}\}$ into adaptive weights $\{w_{\ell}\}$, so the loss focuses on the sub-bands with the strongest subgroup imbalance. $\mathcal{L}_{\mathrm{GCWD}}$ reduces these imbalanced frequency components.
}
\label{fig:gcwd_pipeline}
\end{figure}

GCWD targets unfairness before pooling.
The intuition is that demographic appearance cues and forgery cues may appear at different spatial scales~\cite{qian2020thinking}.
If fairness supervision is applied only after pooling, these signals have already been mixed. 
GCWD instead operates on the spatial feature map $\mathbf{F}_i$ and reduces subgroup imbalance in its wavelet statistics (Figure~\ref{fig:gcwd_pipeline}).

\textbf{Wavelet decomposition.}
For each spatial feature map $\mathbf{F}_i$, we apply a channel-wise 2D Daubechies-4 Discrete Wavelet Transform (DWT)~\cite{daubechies1992wavelets,mallat1989wavelet}. 
DWT decomposes $\mathbf{F}_i$ into $L$ sub-band tensors $\{\mathbf{C}_{i,\ell}\}_{\ell=1}^{L}$, where $L$ is determined by the decomposition depth. 
The decomposition produces one low-frequency approximation band $\mathrm{LL}$ and, at each level, three detail bands $\mathrm{LH}$, $\mathrm{HL}$ and $\mathrm{HH}$, corresponding to high-frequency variations along different spatial directions. 
This lets GCWD measure subgroup imbalance in frequency-localised components rather than on the whole feature map. 
When using two decomposition levels, the resulting bands are $\{\mathrm{LL}_{2},\mathrm{LH}_{2},\mathrm{HL}_{2},\mathrm{HH}_{2},\mathrm{LH}_{1},\mathrm{HL}_{1},\mathrm{HH}_{1}\}$, giving $L=7$. 
We define the energy of sample $i$ in sub-band $\ell$ as $e_{i,\ell}=\|\mathbf{C}_{i,\ell}\|_{F}^{2}$.

\textbf{Group-conditional sub-band discrepancy.}
After decomposing $\mathbf{F}_i$ into sub-bands, GCWD asks whether each sub-band is used similarly across demographic subgroups. 
If a sub-band has very different energy across groups, the model may be encoding subgroup-related appearance in that frequency range. 
We therefore measure the cross-group variation of sub-band energy and use it as the fairness signal.

For a mini-batch $\mathcal{M}$, let $\mathcal{M}_{g}$ denote the samples from subgroup $g$. 
We compute the average energy of band $\ell$ for subgroup $g$ as
$E_{\ell}^{g}=|\mathcal{M}_{g}|^{-1}\sum_{i\in\mathcal{M}_{g}}e_{i,\ell}$, 
and use the cross-subgroup average 
$\bar{E}_{\ell}=G^{-1}\sum_{g=1}^{G}E_{\ell}^{g}$ as the reference energy for this band. 
The subgroup discrepancy $D_{\ell}$ of band $\ell$ is
\begin{equation}
\label{eq:band_discrepancy}
    D_{\ell}
    =
    \sum_{g=1}^{G}
    \left(E_{\ell}^{g}-\bar{E}_{\ell}\right)^{2}.
\end{equation}
A larger $D_{\ell}$ means that band $\ell$ is more subgroup-imbalanced, while a smaller $D_{\ell}$ means that the band is used more consistently across subgroups.

\textbf{Adaptive sub-band weighting.}
A direct choice would be to minimise the sum of discrepancies $\{D_{\ell}\}$ over all bands. 
However, many bands may already be balanced across subgroups, and some high-frequency bands may contain useful forgery evidence. 
GCWD therefore assigns larger weights to bands with stronger subgroup discrepancy and applies weaker supervision to bands that are already subgroup-consistent.

Specifically, we convert the current discrepancy values $\{D_{\ell}\}_{\ell\in\mathcal{A}}$ into adaptive weights $w_{\ell}$ by a temperature-controlled softmax:
\begin{equation}
\label{eq:gcwd}
    w_{\ell}
    =
    \frac{
    \exp(\operatorname{sg}(D_{\ell})/\tau)
    }{
    \sum_{\ell'\in\mathcal{A}}
    \exp(\operatorname{sg}(D_{\ell'})/\tau)
    },
    \qquad
    \mathcal{L}_{\mathrm{GCWD}}
    =
    \sum_{\ell\in\mathcal{A}}w_{\ell}D_{\ell},
\end{equation}
where $\tau$ controls how concentrated the weights are, and $\operatorname{sg}(\cdot)$ denotes stop-gradient. 

Since $D_{\ell}$ measures how unevenly band $\ell$ behaves across subgroups, minimising the weighted sum directly reduces the most subgroup-imbalanced frequency components. 
The stop-gradient makes $w_{\ell}$ act only as an importance score computed from the current batch, preventing the model from trivially changing the weights instead of reducing the discrepancies. 
Thus, GCWD suppresses subgroup-dependent spatial-frequency imbalance by flattening the cross-group energy differences in the most biased bands, without forcing the entire spatial feature map to be group-invariant.

\subsection{Subspace-Localised Mean Alignment (SLMA)}
\label{sec:slma}

GCWD reduces subgroup imbalance in the spatial feature map, but it does not directly control the bias introduced during fine-tuning. 
We do not assume that the frozen principal component $\mathbf{W}_{p}$ is demographic-free. 
As the dominant part of a pre-trained visual representation, $\mathbf{W}_{p}$ inevitably encodes facial appearance, including skin tone, face shape, and gender-correlated cues. 
However, these features belong to the frozen visual prior and are not changed by deepfake fine-tuning. 
The fairness risk we target is different: when the training data is demographically imbalanced, the trainable residual component $\Delta\mathbf{W}$ can absorb subgroup-specific shifts and turn them into shortcuts for real/fake prediction.

SLMA therefore regularises the residual view $\mathbf{z}_{\Delta,i}$ (Eq.~\eqref{eq:zdelta}) rather than the full pooled feature $\mathbf{z}_i$. 
This design constrains the adaptation pathway through which unfairness can be newly introduced during fine-tuning.
Intuitively, among real samples, the residual update should not systematically shift one subgroup away from another; the same should also hold among fake samples. 
Otherwise, the classifier may exploit subgroup-dependent residual shifts as shortcuts for prediction~\cite{geirhos2020shortcut}.
We implement this idea with class-conditional first-moment matching~\cite{madras2018learning,edwards2016censoring,cheng2026fair}, but localise it to the residual view $\mathbf{z}_{\Delta,i}$ rather than the full feature $\mathbf{z}_i$.

For class $y$ and subgroup $g$, let $\mathcal{M}_{y,g}$ denote the mini-batch samples with label $y$ and subgroup $g$. 
We compute the average residual representation $\boldsymbol{\mu}_{y,g}$ of subgroup $g$ within class $y$ and use the class-wise subgroup average $\bar{\boldsymbol{\mu}}_{y}$ as the reference:
\begin{equation}
\label{eq:slma_mu}
    \boldsymbol{\mu}_{y,g}
    =
    \frac{1}{|\mathcal{M}_{y,g}|}
    \sum_{i\in\mathcal{M}_{y,g}}
    \mathbf{z}_{\Delta,i},
    \qquad
    \bar{\boldsymbol{\mu}}_{y}
    =
    \frac{1}{G}
    \sum_{g=1}^{G}
    \boldsymbol{\mu}_{y,g}.
\end{equation}
Here, $\boldsymbol{\mu}_{y,g}$ describes how fine-tuning shifts subgroup $g$ on average within class $y$, while $\bar{\boldsymbol{\mu}}_{y}$ represents the subgroup-neutral residual shift for that class. 
SLMA penalises subgroup-specific deviations from this reference:
\begin{equation}
\label{eq:slma}
    \mathcal{L}_{\mathrm{SLMA}}
    =
    \sum_{y\in\{0,1\}}
    \sum_{g=1}^{G}
    \left\|
    \boldsymbol{\mu}_{y,g}
    -
    \bar{\boldsymbol{\mu}}_{y}
    \right\|_{2}^{2}.
\end{equation}
Conditioning on $y$ is essential.
It aligns subgroups only among samples with the same real/fake label, so the loss reduces demographic shifts in the residual pathway without pulling real and fake representations together.

\subsection{Training Objective and Inference}
\label{sec:training}

FairReL is trained with one detection objective and three auxiliary regularisation terms. 
The binary cross-entropy loss $\mathcal{L}_{\mathrm{CE}}$ trains the real/fake prediction from $\hat{p}_i$, $\mathcal{L}_{\mathrm{orth}}$ (Eq. \ref{eq:orth}) preserves the SVD-based separation between principal component and trainable residual, $\mathcal{L}_{\mathrm{GCWD}}$ (Eq. \ref{eq:gcwd}) reduces subgroup imbalance in the spatial feature map $\mathbf{F}_i$, and $\mathcal{L}_{\mathrm{SLMA}}$ (Eq. \ref{eq:slma}) suppresses subgroup-dependent shifts in the residual view $\mathbf{z}_{\Delta,i}$.  
The full objective is
\begin{equation}
\label{eq:total_loss}
    \mathcal{L}
    =
    \mathcal{L}_{\mathrm{CE}}(\hat{p},y)
    +
    \lambda_{\mathrm{orth}}\mathcal{L}_{\mathrm{orth}}
    +
    \lambda_{G}\mathcal{L}_{\mathrm{GCWD}}
    +
    \lambda_{S}\mathcal{L}_{\mathrm{SLMA}},
\end{equation}
where $\lambda_{\mathrm{orth}}$, $\lambda_G$, and $\lambda_S$ control the auxiliary terms. 
Only the residual weights $\Delta\mathbf{W}$ and the classifier parameters $h_{\phi}$ are updated, while the principal weights $\mathbf{W}_{p}$ remain frozen.

At inference, FairReL reduces to the standard detector $\hat{p}=h_{\phi}\left(f_p\left(B_{p+\Delta}(x)\right)\right)$.
The residual-only path, wavelet transform, fairness losses, and demographic labels are removed, so FairReL adds no inference-time overhead.

\section{Experiments}
\label{sec:experiments}

\subsection{Experimental Setup}
\label{sec:setup_exp}

\noindent\textbf{Datasets.}
We evaluate FairReL under a cross-dataset fairness protocol, where the model is trained on a source dataset and tested on both source-domain and unseen target-domain data.
This setting is important because fairness measured on the training domain does not necessarily transfer to new manipulations, identities, or demographic distributions~\cite{lin2024preserving}.
Following prior work~\cite{lin2024preserving,cheng2026fair}, models are trained on FaceForensics++ (FF++, c23)~\cite{rossler2019faceforensics} and tested on three unseen datasets with different data sources and manipulations: Celeb-DF~\cite{li2020celebdf}, DFD~\cite{dufour2019dfd}, and DFDC~\cite{dolhansky2020dfdc}. Further dataset details and subgroup statistics are provided in the supplementary material.

\noindent\textbf{Demographic annotations.}
We use the demographic annotations released by \citet{lin2024preserving}, which follow the pipeline of prior fairness studies~\cite{ju2024improving,xu2024analyzing}.
We report fairness over the intersectional gender--race partition with six subgroups: Male-White (M-W), Male-Asian (M-A), Male-Black (M-B), Female-White (F-W), Female-Asian (F-A) and Female-Black (F-B).
Demographic labels are not required by the detector at inference.

\noindent\textbf{Detection metrics.}
We report the frame-level Area Under the ROC Curve (AUC) for real/fake separability, and additionally a group-averaged AUC ($\mathrm{AUC}_{\mathrm{avg}}$) defined as the mean of per-subgroup AUCs, which weights every demographic subgroup equally and exposes detection quality on minority subgroups that the overall AUC would otherwise hide.

\noindent\textbf{Fairness metrics.}
We evaluate subgroup fairness mainly through False Positive Rate (FPR) disparity following Ju~\etal~\cite{ju2024improving} and Lin~\etal~\cite{lin2024preserving}, since a false positive wrongly flags a real face as fake and can directly harm innocent users.
We report \emph{Equal FPR} ($\FFPR$), which measures how much each subgroup's FPR deviates from the overall FPR; and \emph{Max Equalized Odds} ($\FMEO$), which captures the worst-case subgroup gap over either FPR or TPR.
Formal definitions are provided in the supplementary material.
Since FPR gaps can be artificially reduced by changing the decision threshold $\tau$ (for classifier $h_{\phi}$ in Eq.~\ref{eq:forward}), we follow Grother~\etal~\cite{grother2019frvt} and fix $\tau$ on the FF++ validation split at source-domain True Positive Rate (TPR) of 0.90 for all FPR-based metrics.

\noindent\textbf{Baseline methods.}
To provide a comprehensive evaluation, we compare FairReL with three categories of baselines.
(i) ERM baselines: Xception~\cite{chollet2017xception} and CLIP-ERM, where CLIP-ERM denotes an ERM classifier trained on CLIP ViT-L/14 features~\cite{radford2021clip}.
(ii) Generalisation-oriented detectors, designed for cross-domain robustness rather than demographic fairness and all built on CLIP ViT-L/14: ForAda~\cite{cui2025forensics}, Effort~\cite{yan2024effort} (which also serves as our SVD-decomposed backbone) and GenD~\cite{yermakov2026deepfake}.
(iii) Fairness-oriented detectors: DAW-FDD~\cite{ju2024improving} and PG-FDD~\cite{lin2024preserving} on Xception, and FairAdapter~\cite{ding2025fairadapter} on CLIP ViT-L/14.
All reproduced baselines and FairReL are evaluated under the same protocol described above.

\noindent\textbf{Implementation details.}
We use CLIP ViT-L/14~\cite{radford2021clip,ojha2023universal} with SVD decomposition on all self-attention projections. 
The residual rank $r$ is set to 1 following~\cite{yan2024effort}. 
Loss weights are selected on the FF++ validation split as $\lambda_{\mathrm{orth}}=0.1$ and $\lambda_G=\lambda_S=0.5$. 
Further details are provided in the supplementary material.

\subsection{Main Results}
\label{sec:sota}
\label{sec:overall_cmp}

\noindent\textbf{Comparison with baseline methods.}
Table~\ref{tab:main_results} evaluates FF++-trained models on three unseen datasets: Celeb-DF, DFD and DFDC.
FairReL achieves the lowest $\FFPR$ and $\FMEO$ on all three unseen datasets while preserving competitive AUC.

\begin{table*}[!htbp]
\begin{center}
\scriptsize
\setlength{\tabcolsep}{2.6pt}
\renewcommand{\arraystretch}{0.9}
\resizebox{\textwidth}{!}{%
\begin{tabular}{lll|cc|cc|cc|cc|cc|cc}
\toprule
\multirow{3}{*}{Type} & \multirow{3}{*}{Method} & \multirow{3}{*}{Backbone}
& \multicolumn{4}{c|}{\textbf{Celeb-DF}}
& \multicolumn{4}{c|}{\textbf{DFD}}
& \multicolumn{4}{c}{\textbf{DFDC}} \\
\cmidrule(lr){4-7}\cmidrule(lr){8-11}\cmidrule(lr){12-15}
& &
& \multicolumn{2}{c|}{Detection (\%)} & \multicolumn{2}{c|}{Fairness (\%)}
& \multicolumn{2}{c|}{Detection (\%)} & \multicolumn{2}{c|}{Fairness (\%)}
& \multicolumn{2}{c|}{Detection (\%)} & \multicolumn{2}{c}{Fairness (\%)} \\
\cmidrule(lr){4-5}\cmidrule(lr){6-7}\cmidrule(lr){8-9}\cmidrule(lr){10-11}\cmidrule(lr){12-13}\cmidrule(lr){14-15}
& &
& AUC$\uparrow$ & AUC$_{\mathrm{avg}}\uparrow$ & $\FFPR\downarrow$ & $\FMEO\downarrow$
& AUC$\uparrow$ & AUC$_{\mathrm{avg}}\uparrow$ & $\FFPR\downarrow$ & $\FMEO\downarrow$
& AUC$\uparrow$ & AUC$_{\mathrm{avg}}\uparrow$ & $\FFPR\downarrow$ & $\FMEO\downarrow$ \\
\midrule
\multirow{2}{*}{ERM}
& Xception~\cite{chollet2017xception} & Xception
& 70.94 & 70.75 & 25.43 & 16.81
& 76.73 & 76.57 & 29.20 & 18.11
& 60.58 & 60.45 & 48.89 & 37.15 \\
& CLIP-ERM~\cite{radford2021clip} & CLIP
& 74.37 & 74.26 & 23.25 & 15.71
& 81.14 & 81.06 & 31.54 & 20.69
& 62.35 & 62.30 & 51.15 & 39.46 \\
\midrule
\multirow{3}{*}{Gen.}
& ForAda~\cite{cui2025forensics} & CLIP
& 80.21 & 80.26 & 20.35 & 12.75
& 85.50 & 85.53 & 27.41 & 23.46
& 64.82 & 64.87 & 49.70 & 38.61 \\
& Effort~\cite{yan2024effort} & CLIP
& 78.52 & 78.58 & 19.61 & 13.44
& 84.22 & 84.26 & 19.37 & 15.08
& 65.08 & 65.14 & 52.04 & 40.31 \\
& GenD~\cite{yermakov2026deepfake} & CLIP
& \textbf{80.85} & \underline{80.92} & 18.76 & 11.07
& \textbf{86.43} & \underline{86.48} & 20.01 & 17.77
& \textbf{66.85} & \underline{66.88} & 50.81 & 41.35 \\
\midrule
\multirow{3}{*}{Fair.}
& DAW-FDD~\cite{ju2024improving} & Xception
& 70.46 & 70.52 & 15.11 & 11.18
& 73.65 & 73.71 & 18.23 & 13.41
& 58.41 & 58.53 & 43.26 & 35.85 \\
& PG-FDD~\cite{lin2024preserving} & Xception
& 75.42 & 75.57 & \underline{12.82} & \underline{9.12}
& 82.32 & 82.47 & 15.67 & 10.30
& 63.70 & 63.79 & 40.56 & 33.97 \\
& FairAdapter~\cite{ding2025fairadapter} & CLIP
& 75.65 & 75.82 & 13.48 & 10.56
& 80.76 & 80.92 & \underline{14.86} & \underline{9.11}
& 62.16 & 62.38 & \underline{39.02} & \underline{33.22} \\
\midrule
\rowcolor{ourrow}
Ours & \textbf{FairReL} & CLIP
& \underline{80.64} & \textbf{81.26} & \textbf{10.12} & \textbf{7.03}
& \underline{86.13} & \textbf{86.78} & \textbf{13.55} & \textbf{8.98}
& \underline{66.34} & \textbf{67.03} & \textbf{38.36} & \textbf{30.02} \\
\bottomrule
\end{tabular}%
}
\end{center}
\caption{
Cross-domain comparison on Celeb-DF, DFD and DFDC after training on FF++.
``Type'' groups methods into ERM baselines, generalisation-oriented detectors (Gen.), and fairness-aware detectors (Fair.).
$\uparrow$/$\downarrow$ indicate higher/lower is better.
The best results are shown in \textbf{bold} and the second-best are \underline{underlined}.
}

\label{tab:cross}
\label{tab:main}
\label{tab:main_results}
\end{table*}

\begin{table*}[!htbp]
\begin{center}
\scriptsize
\setlength{\tabcolsep}{4pt}
\renewcommand{\arraystretch}{1}
\begin{tabular}{ll|cccccc|ccc}
\toprule
& & \multicolumn{6}{c|}{Per-subgroup AUC (\%)$\uparrow$}
& \multicolumn{3}{c}{Overall (\%)} \\
\cmidrule(lr){3-8}\cmidrule(lr){9-11}
Type & Method
& M-W & M-A & M-B & F-W & F-A & F-B
& AUC$\uparrow$ & AUC$_{\mathrm{avg}}\uparrow$ & AUC gap$\downarrow$ \\
\midrule
\multirow{2}{*}{ERM}
& Xception~\cite{chollet2017xception}
& 61.06 & 60.69 & 60.28 & 61.56 & 59.78 & 59.34 & 60.58 & 60.45 & 2.22 \\
& CLIP-ERM~\cite{radford2021clip}
& 62.87 & 62.53 & 62.14 & 63.34 & 61.67 & 61.26 & 62.35 & 62.30 & 2.08 \\
\midrule
\multirow{3}{*}{Gen.}
& ForAda~\cite{cui2025forensics}
& 65.38 & 65.07 & 64.73 & 65.80 & 64.31 & 63.94 & 64.82 & 64.87 & 1.86 \\
& Effort~\cite{yan2024effort}
& 65.64 & 65.34 & 65.00 & 66.06 & 64.59 & 64.22 & 65.08 & 65.14 & 1.84 \\
& GenD~\cite{yermakov2026deepfake}
& \textbf{67.93} & \underline{67.19} & \underline{66.59} & \textbf{67.86} & \underline{65.91} & \underline{65.77} & \textbf{66.85} & \underline{66.88} & 2.16 \\
\midrule
\multirow{3}{*}{Fair.}
& DAW-FDD~\cite{ju2024improving}
& 58.86 & 58.66 & 58.43 & 59.14 & 58.16 & 57.92 & 58.41 & 58.53 & 1.22 \\
& PG-FDD~\cite{lin2024preserving}
& 64.11 & 63.92 & 63.70 & 64.37 & 63.44 & 63.21 & 63.70 & 63.79 & \underline{1.16} \\
& FairAdapter~\cite{ding2025fairadapter}
& 62.73 & 62.52 & 62.28 & 63.02 & 61.99 & 61.74 & 62.16 & 62.38 & 1.28 \\
\midrule
\rowcolor{ourrow}
Ours & \textbf{FairReL}
& \underline{67.35} & \textbf{67.42} & \textbf{67.02} & \underline{67.10} & \textbf{66.33} & \textbf{66.95} & \underline{66.34} & \textbf{67.03} & \textbf{1.09} \\
\bottomrule
\end{tabular}
\end{center}
\caption{
Per-subgroup AUC (\%) on DFDC (trained on FF++).
}
\label{tab:dfdc_subgroup}
\label{tab:subgroup_case}
\label{tab:subgroup_auc}
\end{table*}

FairReL slightly trails the strongest generalisation-oriented detector GenD~\cite{yermakov2026deepfake} in overall AUC but achieves higher $\mathrm{AUC}_{\mathrm{avg}}$ and lower fairness disparities.
Cross-domain robustness alone thus does not remove subgroup-sensitive structure: forgery-relevant features still carry subgroup imbalance.
FairReL therefore localises fairness supervision to these two sources of imbalance.

Against fairness-oriented baselines such as DAW-FDD~\cite{ju2024improving} and PG-FDD~\cite{lin2024preserving}, FairReL further lowers cross-domain fairness disparities: rather than regularising only the prediction layer, it intervenes at these two localised sources of bias, which explains its stronger transfer under distribution shift.

\noindent\textbf{Per-subgroup analysis.}
\label{sec:subgroup}
In Table~\ref{tab:subgroup_auc}, FairReL achieves the highest $\mathrm{AUC}_{\mathrm{avg}}$ and the smallest AUC gap, while GenD~\cite{yermakov2026deepfake} remains higher on the two White subgroups.
This indicates that FairReL transfers detection quality more evenly across subgroups under domain shift, while GenD's small overall-AUC lead is concentrated in the majority White subgroups rather than uniformly across all groups.

\begin{figure*}[!htbp]
\centering
\includegraphics[width=\textwidth]{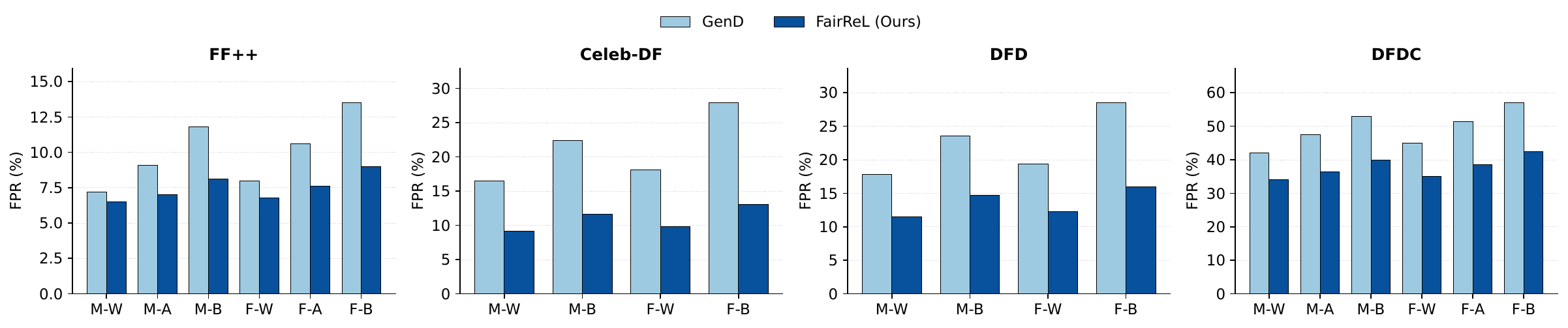}
\vspace{-4mm}
\caption{
Per-subgroup FPR (\%) of GenD~\cite{yermakov2026deepfake} and FairReL.
}
\label{fig:subgroup_fpr}
\end{figure*}

We further examine per-subgroup FPR at the threshold fixed on the FF++ validation split at TPR=0.90.
Figure~\ref{fig:subgroup_fpr} shows that FairReL lowers worst-subgroup FPR and narrows the best--worst gap across all four datasets.
This improvement is most visible on minority Female and Black subgroups, where GenD produces concentrated false positives.
These results suggest that FairReL reduces subgroup-dependent decision shortcuts rather than merely improving the average detector.
FairReL does this by targeting two sites: GCWD weakens subgroup-imbalanced spatial frequencies before pooling, while SLMA suppresses adaptation-induced subgroup shifts in the residual, which we verify directly in the next section.

\subsection{Mechanistic Analysis of FairReL}
\label{sec:understanding}
In this section, we examine the sources of subgroup unfairness that motivate FairReL. 
We analyse whether unfairness is concentrated in spatial-frequency structure, motivating GCWD, and in the fine-tuning residual representation, motivating SLMA. 
All diagnostics use DFDC~\cite{dolhansky2020dfdc}, with results on other targets provided in the appendix.

\noindent\textbf{Frequency-wise analysis.}
We first locate subgroup-dependent variation across the wavelet sub-bands of the spatial feature map. 
GCWD computes the subgroup discrepancy $D_\ell$ for each band $\ell$ (Eq.~\ref{eq:band_discrepancy}).
A larger $D_\ell$ means stronger subgroup-dependent variation in band $\ell$, so plotting $D_\ell$ across bands reveals where demographic imbalance is most exposed.

\begin{figure}[!htbp]
\centering
\begin{minipage}{0.45\columnwidth}
\centering
\includegraphics[width=\linewidth]{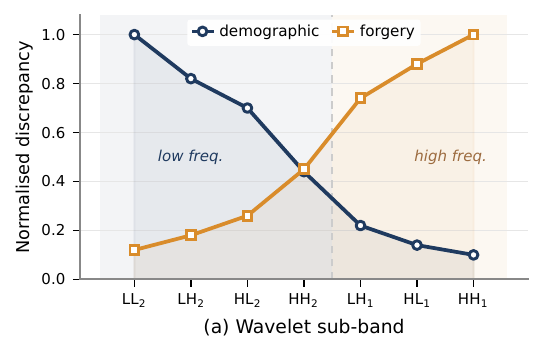}
\end{minipage}
\hspace{0.01\columnwidth}
\begin{minipage}{0.45\columnwidth}
\centering
\includegraphics[width=\linewidth]{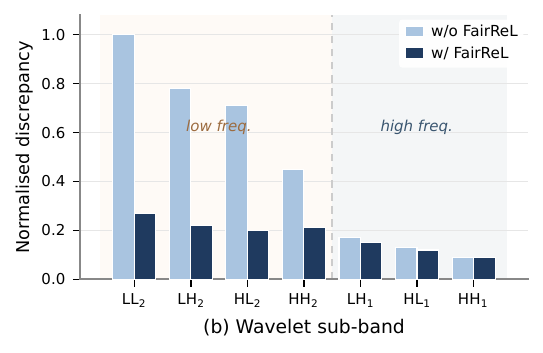}
\end{minipage}
\vspace{3mm}
\caption{
Frequency-wise diagnostics of subgroup imbalance and forgery separability.
(a) Sub-band diagnostic statistics.
(b) Sub-band discrepancy before and after FairReL.
}
\label{fig:subband_discrepancy}
\end{figure}

Figure~\ref{fig:subband_discrepancy}(a) shows that the normalised $D_\ell$ peaks in the coarse, low-frequency bands, where global appearance statistics such as skin tone, illumination, and colour distribution are more visible. 
In contrast, the normalised real--fake energy gap is stronger in finer bands, which better capture local texture, boundary and synthesis artefacts. 
This separation explains why GCWD regularises sub-band discrepancies rather than the whole feature map. 
It identifies frequency bands with large subgroup imbalance and applies stronger decorrelation to those bands, while leaving more balanced or forgery-discriminative bands less constrained. 
Consistently, Figure~\ref{fig:subband_discrepancy}(b) shows that FairReL mainly reduces the low-frequency subgroup discrepancy, where demographic imbalance is strongest, while largely preserving the higher-frequency bands that carry stronger real--fake separation.

\noindent\textbf{Residual analysis.}
We next examine whether fine-tuning-induced subgroup shifts are concentrated in the residual pathway. 
To test this, we compare a principal-only view ($\Delta\mathbf{W}\!\leftarrow\!\mathbf{0}$), where the residual component is removed, and a residual-only view ($\mathbf{W}_p\!\leftarrow\!\mathbf{0}$), where the frozen principal component is removed.
For each view, we measure the subgroup mean shift (Eq.\ref{eq:slma_mu}) by computing the distance from the average feature of each subgroup to the overall average.
If samples from one subgroup have a different average representation from other subgroups, the classifier may learn from this unfair shortcut.

\begin{figure*}[!htbp]
\centering
\includegraphics[width=1\textwidth]{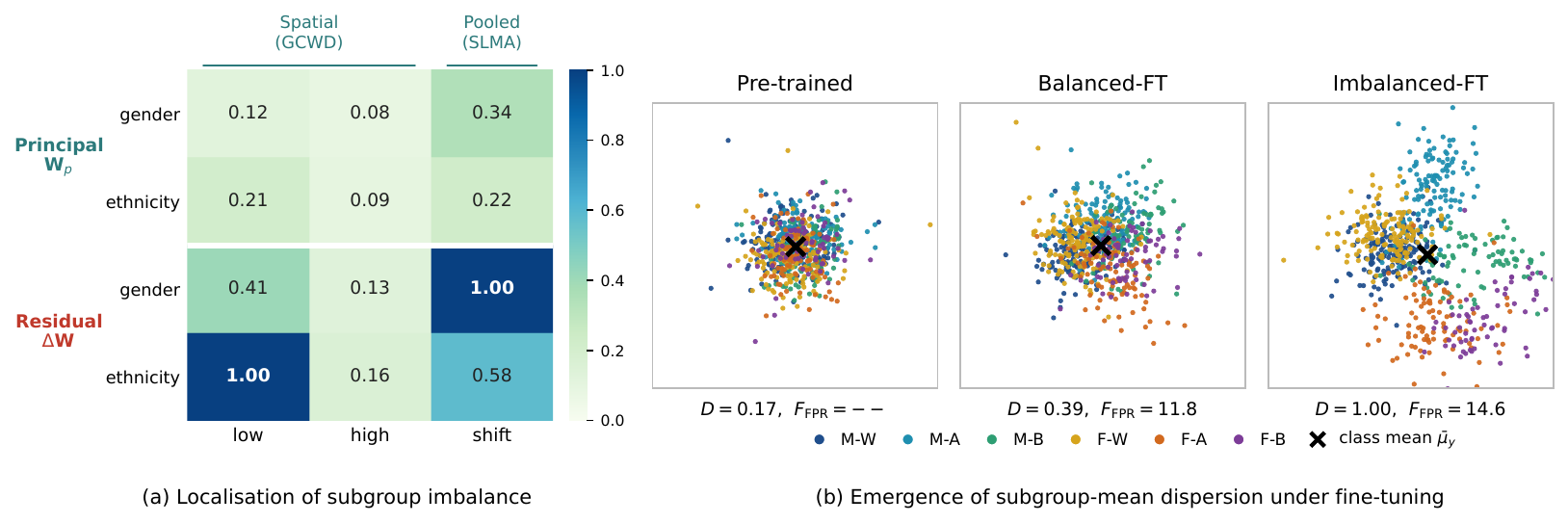}
\vspace{-3mm}
\caption{
Mechanistic diagnostics on DFDC.
(a) Principal-vs-residual comparison of subgroup imbalance across spatial-frequency and pooled-feature statistics.
(b) PCA visualisation of residual subgroup centroids after subgroup-balanced/imbalanced fine-tuning.
}
\label{fig:mechanism}
\end{figure*}

Figure~\ref{fig:mechanism}(a) separates this analysis by attribute and bias type.
The low-frequency discrepancy is stronger than high-frequency discrepancy, especially for ethnicity, supporting the spatial-frequency motivation of GCWD.
More importantly for SLMA, the pooled mean shift is consistently larger in the residual view than in the principal view, with the strongest effect appearing for gender.
This shows that fine-tuning concentrates subgroup bias as subgroup mean shifts in the trainable residual pathway.
Therefore, existing fairness-aware methods, which regularise the full pooled feature, do not distinguish the residual shift from the principal representation that supports real/fake discrimination.
SLMA avoids this problem by aligning subgroup means only in the residual representation and only within each real/fake class, directly reducing adaptation-induced bias while preserving the full detection feature.

\noindent\textbf{Feature dynamics.}
We finally examine how subgroup shifts emerge during fine-tuning.
Figure~\ref{fig:mechanism}(b) visualises the residual feature $\mathbf{z}_\Delta$ with subgroup centroids under three stages: before fine-tuning, balanced fine-tuning, and imbalanced fine-tuning.
Before fine-tuning, the subgroup centroids are relatively compact.
Balanced fine-tuning introduces only moderate separation, whereas imbalanced fine-tuning spreads the subgroup centroids much further apart.
Demographic imbalance in the training data is thus injected into the residual representation during adaptation, precisely the bias source targeted by SLMA.

\subsection{Ablation Studies}
\label{sec:ablation}

\noindent\textbf{Necessity of subspace decomposition.}
We first test whether the SVD residual split is necessary.
Table~\ref{tab:svd_localisation} compares FairReL with a variant that fully fine-tunes CLIP and applies GCWD with SLMA on the full pooled feature $\mathbf{z}$.
The full-feature variant is competitive yet consistently worse than FairReL in both AUC and $\FFPR$.
This supports localising fairness supervision to the trainable residual.

\begin{table}[!htbp]
\begin{center}
\footnotesize
\setlength{\tabcolsep}{5pt}
\renewcommand{\arraystretch}{1.0}
\begin{tabular}{lcc|cc|cc|cc}
\toprule
& \multicolumn{2}{c|}{\textbf{FF++}}
& \multicolumn{2}{c|}{\textbf{Celeb-DF}}
& \multicolumn{2}{c|}{\textbf{DFD}}
& \multicolumn{2}{c}{\textbf{DFDC}} \\
Variant
& AUC$\uparrow$ & $\FFPR\downarrow$
& AUC$\uparrow$ & $\FFPR\downarrow$
& AUC$\uparrow$ & $\FFPR\downarrow$
& AUC$\uparrow$ & $\FFPR\downarrow$ \\
\midrule
w/o subspace decomposition
& 95.37 & 12.83 & 79.12 & 11.87 & 84.91 & 15.26 & 64.79 & 40.47 \\
\midrule
\rowcolor{ourrow}\textbf{FairReL (w/ decomposition)}
& \textbf{96.14} & \textbf{10.97} & \textbf{80.64} & \textbf{10.12} & \textbf{86.13} & \textbf{13.55} & \textbf{66.34} & \textbf{38.36} \\
\bottomrule
\end{tabular}
\end{center}
\caption{
Effect of subspace decomposition. Without decomposition, CLIP is fully fine-tuned; GCWD acts on $\mathbf{F}$ and SLMA is applied to the full pooled feature $\mathbf{z}$.
}
\label{tab:svd_localisation}
\end{table}

\begin{table}[!htbp]
\centering
\tiny
\setlength{\tabcolsep}{2.1pt}
\renewcommand{\arraystretch}{0.88}
\resizebox{\columnwidth}{!}{%
\begin{tabular}{lccc|cc|cc|cc|cc}
\toprule
\multirow{2}{*}{Variant}
& \multicolumn{3}{c|}{Setting}
& \multicolumn{2}{c|}{FF++}
& \multicolumn{2}{c|}{Celeb-DF}
& \multicolumn{2}{c|}{DFD}
& \multicolumn{2}{c}{DFDC} \\
\cmidrule(lr){2-4}\cmidrule(lr){5-6}\cmidrule(lr){7-8}\cmidrule(lr){9-10}\cmidrule(lr){11-12}
& GCWD & SLMA & Cond.
& AUC$\uparrow$ & $\FFPR\downarrow$
& AUC$\uparrow$ & $\FFPR\downarrow$
& AUC$\uparrow$ & $\FFPR\downarrow$
& AUC$\uparrow$ & $\FFPR\downarrow$ \\
\midrule
SVD baseline~\cite{yan2024effort}
& -- & -- & --
& 94.83 & 14.57
& 78.52 & 19.61
& 84.22 & 19.37
& 65.08 & 52.04 \\
GCWD only
& \checkmark & -- & --
& 95.34 & 12.68
& 79.43 & 14.87
& 85.04 & 16.37
& 65.43 & 44.76 \\
SLMA only
& -- & $\mathbf{z}_\Delta$ & \checkmark
& 95.12 & 13.04
& 79.08 & 15.31
& 84.77 & 16.83
& 65.24 & 45.58 \\
\midrule
GCWD + SLMA on $\mathbf{z}$
& \checkmark & $\mathbf{z}$ & \checkmark
& 95.62 & 11.88
& 79.73 & 12.39
& 85.24 & 15.12
& 65.72 & 41.19 \\
GCWD + SLMA on $\mathbf{z}_\Delta$
& \checkmark & $\mathbf{z}_\Delta$ & --
& 93.18 & 11.43
& 77.28 & 11.64
& 83.03 & 14.37
& 62.54 & 40.08 \\
\midrule
\rowcolor{ourrow}\textbf{FairReL}
& \checkmark & $\mathbf{z}_\Delta$ & \checkmark
& \textbf{96.14} & \textbf{10.97}
& \textbf{80.64} & \textbf{10.12}
& \textbf{86.13} & \textbf{13.55}
& \textbf{66.34} & \textbf{38.36} \\
\bottomrule
\end{tabular}}
\caption{
Unified ablation of GCWD, SLMA and class conditioning across four datasets.
``Cond.'' indicates whether SLMA aligns subgroup means within each real/fake class.
}
\label{tab:abl_components}
\end{table}

\begin{table}[!htbp]
\begin{center}
\scriptsize
\setlength{\tabcolsep}{2pt}
\renewcommand{\arraystretch}{0.9}
\resizebox{\columnwidth}{!}{%
\begin{tabular}{lccc|ccc|ccc}
\toprule
& \multicolumn{3}{c|}{\textbf{Celeb-DF}}
& \multicolumn{3}{c|}{\textbf{DFD}}
& \multicolumn{3}{c}{\textbf{DFDC}} \\
Method on the same backbone
& AUC$\uparrow$ & $\FFPR\downarrow$ & $\FMEO\downarrow$
& AUC$\uparrow$ & $\FFPR\downarrow$ & $\FMEO\downarrow$
& AUC$\uparrow$ & $\FFPR\downarrow$ & $\FMEO\downarrow$ \\
\midrule
Subspace-decomposed baseline~\cite{yan2024effort}
& 78.52 & 19.61 & 13.44
& 84.22 & 19.37 & 15.08
& 65.08 & 52.04 & 40.31 \\
\;\;+ prediction-level fairness loss~\cite{madras2018learning}
& 78.43 & 18.92 & 13.08
& 84.11 & 18.57 & 14.73
& 64.91 & 50.82 & 39.58 \\
\;\;+ adversarial classifier (GRL)~\cite{ganin2016dann,zhang2018mitigating}
& 77.88 & 18.23 & 12.64
& 83.62 & 18.08 & 14.12
& 64.23 & 49.71 & 38.87 \\
\;\;+ full-feature mean alignment
& 78.82 & 18.03 & 12.38
& 84.43 & 17.82 & 13.76
& 66.02 & 49.24 & 38.41 \\
\midrule
\rowcolor{ourrow}\textbf{FairReL (Ours)}
& \textbf{80.64} & \textbf{10.12} & \textbf{7.03}
& \textbf{86.13} & \textbf{13.55} & \textbf{8.98}
& \textbf{66.34} & \textbf{38.36} & \textbf{30.02} \\
\bottomrule
\end{tabular}}
\end{center}
\caption{
Comparison of alternative fairness interventions on the same subspace-decomposed backbone under cross-domain evaluation.
}
\label{tab:abl_external}
\end{table}

\noindent\textbf{Loss and alignment ablation.}
\label{sec:abl_components}
Table~\ref{tab:abl_components} jointly ablates GCWD, SLMA and the alignment design.
GCWD and SLMA each improve fairness alone, and their combination gives the best overall balance between AUC and $\FFPR$.
Applying SLMA to $\mathbf{z}_\Delta$ outperforms aligning the full $\mathbf{z}$, and removing class conditioning sharply reduces AUC. Table~\ref{tab:abl_external} further compares against external fairness objectives.

\noindent\textbf{Alternative Fairness Objectives.}
\label{sec:abl_external}
These families share the same subspace-decomposed backbone: \emph{(i)} a prediction-level fairness loss~\cite{madras2018learning,hardt2016equality}; \emph{(ii)} an adversarial demographic classifier with gradient-reversal~\cite{ganin2016dann,zhang2018mitigating,edwards2016censoring}; and \emph{(iii)} full-feature mean alignment, which applies first-moment matching on the full pooled $\mathbf{z}$.
The external objectives provide modest fairness improvements, but their gains are smaller than FairReL and often come with reduced cross-domain AUC. The strongest baseline, full-feature mean alignment, still trails FairReL on all cross-domain fairness metrics: fairness supervision is more effective on the trainable residual than on the full pooled representation.

\subsection{Visualisation Results}
\label{sec:visualization}

\begin{figure*}[!htbp]
\centering
\includegraphics[width=0.99\textwidth]{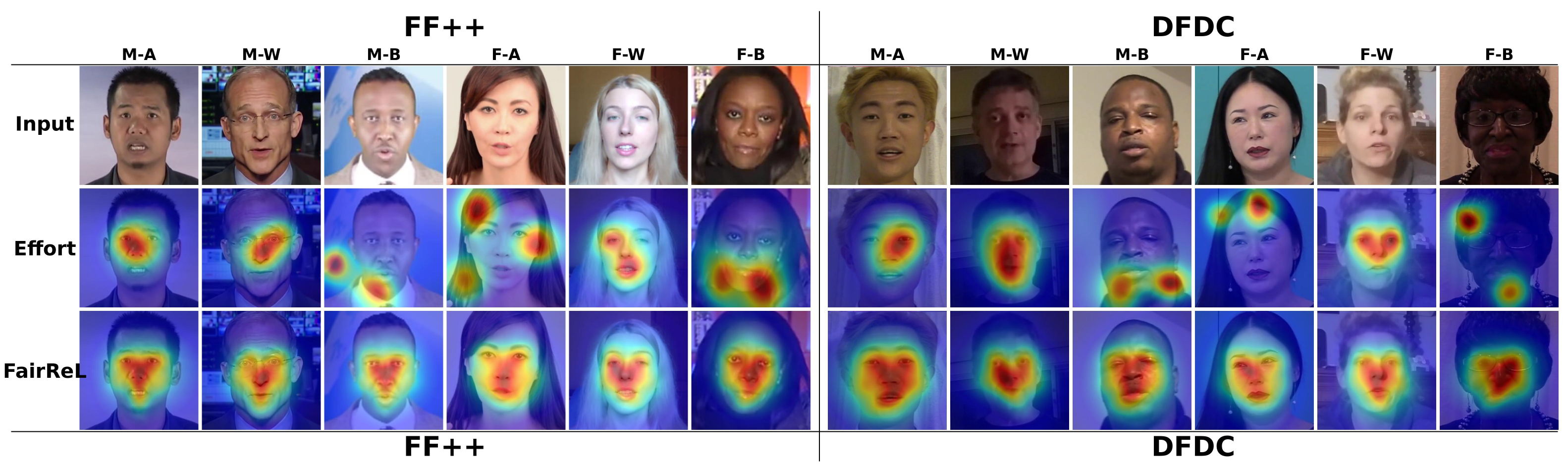}
\caption{
Grad-CAM attribution across the six subgroups on FF++ (left) and DFDC (right).
}
\label{fig:gradcam}
\end{figure*}

Figure~\ref{fig:gradcam} presents Grad-CAM~\cite{selvaraju2017gradcam} visualisations for one test image from each gender--race subgroup on FF++~\cite{rossler2019faceforensics} and DFDC~\cite{dolhansky2020dfdc}.
Compared with Effort~\cite{yan2024effort}, which we use as the SVD-decomposed backbone, FairReL produces more spatially consistent activations across subgroups, concentrated on central facial regions rather than subgroup-correlated areas such as hair or background.
These examples suggest that FairReL yields less subgroup-dependent attention while preserving facial evidence for classification.

\section{Conclusion}
\label{sec:conclusion}

We presented \textbf{FairReL}, a fairness-aware representation-learning framework for generalisable deepfake detection.
Unlike existing approaches that regularise only the final prediction or the pooled representation as a whole, FairReL targets two sources of subgroup bias: spectral imbalance in spatial features and adaptation-induced shifts in the trainable residual subspace.
GCWD reduces subgroup-imbalanced wavelet statistics before pooling, while SLMA aligns class-conditional subgroup means in the residual representation of an SVD-decomposed backbone.
Experiments on FF++, Celeb-DF, DFD and DFDC show that FairReL reduces cross-domain fairness disparities while preserving competitive detection performance.
Ablations and representation analyses further confirm that the two objectives are complementary.
Overall, FairReL improves fairness and cross-dataset generalisation on standard face-manipulation deepfake benchmarks with available demographic annotations.

\section*{Acknowledgements}
This work was supported by the China Scholarship Council--Warwick Joint Scholarship.

\bibliography{BMVCSubmission}

\begin{thebibliography}{50}
\providecommand{\natexlab}[1]{#1}
\providecommand{\url}[1]{\texttt{#1}}
\expandafter\ifx\csname urlstyle\endcsname\relax
  \providecommand{\doi}[1]{doi: #1}\else
  \providecommand{\doi}{doi: \begingroup \urlstyle{rm}\Url}\fi

\bibitem[Agarwal and Ratha(2024)]{agarwal2024deepfake}
Akshay Agarwal and Nalini Ratha.
\newblock Deepfake: Classifiers, fairness, and demographically robust
  algorithm.
\newblock In \emph{2024 IEEE 18th International Conference on Automatic Face
  and Gesture Recognition}, pages 1--9. IEEE, 2024.

\bibitem[Buolamwini and Gebru(2018)]{buolamwini2018gender}
Joy Buolamwini and Timnit Gebru.
\newblock Gender shades: Intersectional accuracy disparities in commercial
  gender classification.
\newblock In \emph{Conference on Fairness, Accountability and Transparency},
  pages 77--91, 2018.

\bibitem[Cheng et~al.(2025)Cheng, Liu, Guo, et~al.]{cheng2026fair}
H.~Cheng, M.~H. Liu, Y.~Guo, et~al.
\newblock Fair deepfake detectors can generalize.
\newblock \emph{Advances in Neural Information Processing Systems},
  38:\penalty0 52596--52621, 2025.

\bibitem[Chollet(2017)]{chollet2017xception}
Fran\c{c}ois Chollet.
\newblock Xception: Deep learning with depthwise separable convolutions.
\newblock In \emph{Proceedings of the IEEE Conference on Computer Vision and
  Pattern Recognition}, pages 1251--1258, 2017.

\bibitem[Cui et~al.(2025)Cui, Li, Luo, Zhou, and Dong]{cui2025forensics}
Xinjie Cui, Yuezun Li, Ao~Luo, Jiaran Zhou, and Junyu Dong.
\newblock Forensics adapter: Adapting clip for generalizable face forgery
  detection.
\newblock In \emph{Proceedings of the Computer Vision and Pattern Recognition
  Conference (CVPR)}, pages 19207--19217, 2025.

\bibitem[Daubechies(1992)]{daubechies1992wavelets}
Ingrid Daubechies.
\newblock \emph{Ten Lectures on Wavelets}, volume~61 of \emph{CBMS-NSF Regional
  Conference Series in Applied Mathematics}.
\newblock SIAM, 1992.

\bibitem[Dhariwal and Nichol(2021)]{dhariwal2021diffusion}
Prafulla Dhariwal and Alexander Nichol.
\newblock Diffusion models beat gans on image synthesis.
\newblock \emph{Advances in neural information processing systems},
  34:\penalty0 8780--8794, 2021.

\bibitem[Ding et~al.(2025)Ding, Zhang, He, and Xu]{ding2025fairadapter}
Feng Ding, Jun Zhang, Xinan He, and Jianfeng Xu.
\newblock Fairadapter: Detecting ai-generated images with improved fairness.
\newblock In \emph{ICASSP 2025 -- 2025 IEEE International Conference on
  Acoustics, Speech and Signal Processing (ICASSP)}, pages 1--5. IEEE, 2025.

\bibitem[Dolhansky et~al.(2020)Dolhansky, Bitton, Pflaum, Lu, Howes, Wang, and
  Ferrer]{dolhansky2020dfdc}
Brian Dolhansky, Joanna Bitton, Ben Pflaum, Jikuo Lu, Russ Howes, Menglin Wang,
  and Cristian~Canton Ferrer.
\newblock The deepfake detection challenge dataset.
\newblock In \emph{arXiv preprint arXiv:2006.07397}, 2020.

\bibitem[Dufour and Gully(2019)]{dufour2019dfd}
Nick Dufour and Andrew Gully.
\newblock Contributing data to deepfake detection research.
\newblock Google AI Blog, 2019.
\newblock URL
  \url{https://research.google/blog/contributing-data-to-deepfake-detection-research/}.
\newblock Google/Jigsaw DeepFake Detection Dataset.

\bibitem[Edwards and Storkey(2016)]{edwards2016censoring}
Harrison Edwards and Amos Storkey.
\newblock Censoring representations with an adversary.
\newblock In \emph{International Conference on Learning Representations}, 2016.

\bibitem[Ezeakunne et~al.(2024)Ezeakunne, Eze, and Liu]{ezeakunne2024data}
Uchenna Ezeakunne, Chigozie Eze, and Xiaoming Liu.
\newblock Data-driven fairness generalization for deepfake detection.
\newblock \emph{arXiv preprint arXiv:2412.16428}, 2024.

\bibitem[Ganin et~al.(2016)Ganin, Ustinova, Ajakan, Germain, Larochelle,
  Laviolette, Marchand, and Lempitsky]{ganin2016dann}
Yaroslav Ganin, Evgeniya Ustinova, Hana Ajakan, Pascal Germain, Hugo
  Larochelle, Fran\c{c}ois Laviolette, Mario Marchand, and Victor Lempitsky.
\newblock Domain-adversarial training of neural networks.
\newblock \emph{Journal of Machine Learning Research}, 17\penalty0
  (59):\penalty0 1--35, 2016.

\bibitem[Geirhos et~al.(2020)Geirhos, Jacobsen, Michaelis, Zemel, Brendel,
  Bethge, and Wichmann]{geirhos2020shortcut}
Robert Geirhos, J{\"o}rn-Henrik Jacobsen, Claudio Michaelis, Richard Zemel,
  Wieland Brendel, Matthias Bethge, and Felix~A. Wichmann.
\newblock Shortcut learning in deep neural networks.
\newblock \emph{Nature Machine Intelligence}, 2\penalty0 (11):\penalty0
  665--673, 2020.

\bibitem[Goodfellow et~al.(2014)Goodfellow, Pouget-Abadie, Mirza, Xu,
  Warde-Farley, Ozair, Courville, and Bengio]{goodfellow2014generative}
Ian~J Goodfellow, Jean Pouget-Abadie, Mehdi Mirza, Bing Xu, David Warde-Farley,
  Sherjil Ozair, Aaron Courville, and Yoshua Bengio.
\newblock Generative adversarial nets.
\newblock \emph{Advances in neural information processing systems}, 27, 2014.

\bibitem[Grother et~al.(2019)Grother, Ngan, and Hanaoka]{grother2019frvt}
Patrick Grother, Mei Ngan, and Kayee Hanaoka.
\newblock Face recognition vendor test part 3: Demographic effects.
\newblock Technical report, National Institute of Standards and Technology,
  2019.

\bibitem[G{"u}era and Delp(2018)]{guera2018temporal}
David G{"u}era and Edward~J. Delp.
\newblock Deepfake video detection using recurrent neural networks.
\newblock In \emph{IEEE International Conference on Advanced Video and Signal
  Based Surveillance}, pages 1--6, 2018.

\bibitem[Hardt et~al.(2016)Hardt, Price, and Srebro]{hardt2016equality}
Moritz Hardt, Eric Price, and Nathan Srebro.
\newblock Equality of opportunity in supervised learning.
\newblock In \emph{Advances in Neural Information Processing Systems
  (NeurIPS)}, pages 3315--3323, 2016.

\bibitem[Ju et~al.(2024)Ju, Hu, Jia, Chen, and Lyu]{ju2024improving}
Yan Ju, Shu Hu, Shan Jia, George~H. Chen, and Siwei Lyu.
\newblock Improving fairness in deepfake detection.
\newblock In \emph{Proceedings of the IEEE/CVF Winter Conference on
  Applications of Computer Vision}, pages 4655--4665, 2024.

\bibitem[Karras et~al.(2019)Karras, Laine, and Aila]{karras2019style}
Tero Karras, Samuli Laine, and Timo Aila.
\newblock A style-based generator architecture for generative adversarial
  networks.
\newblock In \emph{Proceedings of the IEEE/CVF conference on computer vision
  and pattern recognition}, pages 4401--4410, 2019.

\bibitem[Karras et~al.(2020)Karras, Laine, Aittala, Hellsten, Lehtinen, and
  Aila]{karras2020analyzing}
Tero Karras, Samuli Laine, Miika Aittala, Janne Hellsten, Jaakko Lehtinen, and
  Timo Aila.
\newblock Analyzing and improving the image quality of stylegan.
\newblock In \emph{Proceedings of the IEEE/CVF conference on computer vision
  and pattern recognition}, pages 8110--8119, 2020.

\bibitem[Kingma and Welling(2013)]{kingma2013auto}
Diederik~P Kingma and Max Welling.
\newblock Auto-encoding variational bayes.
\newblock \emph{arXiv preprint arXiv:1312.6114}, 2013.

\bibitem[Li et~al.(2020{\natexlab{a}})Li, Bao, Zhang, Yang, Chen, Wen, and
  Guo]{li2020face}
Lingzhi Li, Jianmin Bao, Ting Zhang, Hao Yang, Dong Chen, Fang Wen, and Baining
  Guo.
\newblock Face x-ray for more general face forgery detection.
\newblock In \emph{Proceedings of the IEEE/CVF conference on computer vision
  and pattern recognition}, pages 5001--5010, 2020{\natexlab{a}}.

\bibitem[Li and Lyu(2019)]{li2018warping}
Yuezun Li and Siwei Lyu.
\newblock Exposing deepfake videos by detecting face warping artifacts.
\newblock In \emph{IEEE Conference on Computer Vision and Pattern Recognition
  Workshops}, pages 46--52, 2019.

\bibitem[Li et~al.(2020{\natexlab{b}})Li, Yang, Sun, Qi, and
  Lyu]{li2020celebdf}
Yuezun Li, Xin Yang, Pu~Sun, Honggang Qi, and Siwei Lyu.
\newblock Celeb-df: A large-scale challenging dataset for deepfake forensics.
\newblock In \emph{Proceedings of the IEEE/CVF Conference on Computer Vision
  and Pattern Recognition}, pages 3207--3216, 2020{\natexlab{b}}.

\bibitem[Lin et~al.(2024)Lin, He, Ju, Wang, Ding, and Hu]{lin2024preserving}
Li~Lin, Xinan He, Yan Ju, Xin Wang, Feng Ding, and Shu Hu.
\newblock Preserving fairness generalization in deepfake detection.
\newblock In \emph{Proceedings of the IEEE/CVF Conference on Computer Vision
  and Pattern Recognition}, pages 16815--16825, 2024.

\bibitem[Loshchilov and Hutter(2019)]{loshchilov2019decoupled}
Ilya Loshchilov and Frank Hutter.
\newblock Decoupled weight decay regularization.
\newblock In \emph{International Conference on Learning Representations
  (ICLR)}, 2019.

\bibitem[Madras et~al.(2018)Madras, Creager, Pitassi, and
  Zemel]{madras2018learning}
David Madras, Elliot Creager, Toniann Pitassi, and Richard Zemel.
\newblock Learning adversarially fair and transferable representations.
\newblock In \emph{Proceedings of the International Conference on Machine
  Learning (ICML)}, pages 3384--3393, 2018.

\bibitem[Mallat(1989)]{mallat1989wavelet}
Stephane~G. Mallat.
\newblock A theory for multiresolution signal decomposition: The wavelet
  representation.
\newblock \emph{IEEE Transactions on Pattern Analysis and Machine
  Intelligence}, 11\penalty0 (7):\penalty0 674--693, 1989.

\bibitem[Masood et~al.(2023)Masood, Nawaz, Malik, Javed, Irtaza, and
  Malik]{masood2023deepfakes}
Momina Masood, Mariam Nawaz, Khalid~Mahmood Malik, Ali Javed, Aun Irtaza, and
  Hafiz Malik.
\newblock Deepfakes generation and detection: state-of-the-art, open
  challenges, countermeasures, and way forward: Deepfakes generation and
  detection: state-of-the-art, open challenges, countermeasures, and way
  forward.
\newblock \emph{Applied intelligence}, 53\penalty0 (4):\penalty0 3974--4026,
  2023.

\bibitem[Nadimpalli and Rattani(2022)]{nadimpalli2022gbdf}
Akash~Varma Nadimpalli and Ajita Rattani.
\newblock Gbdf: Gender balanced deepfake dataset towards fair deepfake
  detection.
\newblock In \emph{International Conference on Pattern Recognition}, pages
  320--337. Springer, 2022.

\bibitem[Ojha et~al.(2023)Ojha, Li, and Lee]{ojha2023universal}
Utkarsh Ojha, Yuheng Li, and Yong~Jae Lee.
\newblock Towards universal fake image detectors that generalize across
  generative models.
\newblock In \emph{Proceedings of the IEEE/CVF Conference on Computer Vision
  and Pattern Recognition}, pages 24480--24489, 2023.

\bibitem[Pei et~al.(2026)Pei, Zhang, Hu, Zhang, Wang, Wu, Zhai, Yang, and
  Tao]{pei2026deepfake}
Gan Pei, Jiangning Zhang, Menghan Hu, Zhenyu Zhang, Chengjie Wang, Yunsheng Wu,
  Guangtao Zhai, Jian Yang, and Dacheng Tao.
\newblock Deepfake generation and detection: A benchmark and survey.
\newblock \emph{ACM Computing Surveys}, 58\penalty0 (11):\penalty0 1--41, 2026.

\bibitem[Pu et~al.(2022)Pu, Kuan, Lim, et~al.]{pu2022fairness}
Mengxi Pu, Ming~Yang Kuan, Nicholas~Tze Lim, et~al.
\newblock Fairness evaluation in deepfake detection models using metamorphic
  testing.
\newblock In \emph{Proceedings of the 7th International Workshop on Metamorphic
  Testing}, pages 7--14, 2022.

\bibitem[Qian et~al.(2020)Qian, Yin, Sheng, Chen, and Shao]{qian2020thinking}
Yuyang Qian, Guojun Yin, Lu~Sheng, Zixuan Chen, and Jing Shao.
\newblock Thinking in frequency: Face forgery detection by mining
  frequency-aware clues.
\newblock In \emph{European conference on computer vision}, pages 86--103.
  Springer, 2020.

\bibitem[Radford et~al.(2021)Radford, Kim, Hallacy, Ramesh, Goh, Agarwal,
  Sastry, Askell, Mishkin, Clark, Krueger, and Sutskever]{radford2021clip}
Alec Radford, Jong~Wook Kim, Chris Hallacy, Aditya Ramesh, Gabriel Goh,
  Sandhini Agarwal, Girish Sastry, Amanda Askell, Pamela Mishkin, Jack Clark,
  Gretchen Krueger, and Ilya Sutskever.
\newblock Learning transferable visual models from natural language
  supervision.
\newblock In \emph{Proceedings of the International Conference on Machine
  Learning (ICML)}, pages 8748--8763, 2021.

\bibitem[Ravfogel et~al.(2020)Ravfogel, Elazar, Gonen, Twiton, and
  Goldberg]{ravfogel2020null}
Shauli Ravfogel, Yanai Elazar, Hila Gonen, Michael Twiton, and Yoav Goldberg.
\newblock Null it out: Guarding protected attributes by iterative nullspace
  projection.
\newblock In \emph{Proceedings of the 58th Annual Meeting of the Association
  for Computational Linguistics (ACL)}, pages 7237--7256, 2020.

\bibitem[Ravfogel et~al.(2022)Ravfogel, Twiton, Goldberg, and
  Cotterell]{ravfogel2022linear}
Shauli Ravfogel, Michael Twiton, Yoav Goldberg, and Ryan Cotterell.
\newblock Linear adversarial concept erasure.
\newblock In \emph{Proceedings of the International Conference on Machine
  Learning (ICML)}, pages 18400--18421, 2022.

\bibitem[R{"o}ssler et~al.(2019)R{"o}ssler, Cozzolino, Verdoliva, Riess, Thies,
  and Nie{\ss}ner]{rossler2019faceforensics}
Andreas R{"o}ssler, Davide Cozzolino, Luisa Verdoliva, Christian Riess, Justus
  Thies, and Matthias Nie{\ss}ner.
\newblock Faceforensics++: Learning to detect manipulated facial images.
\newblock In \emph{Proceedings of the IEEE/CVF International Conference on
  Computer Vision}, pages 1--11, 2019.

\bibitem[Selvaraju et~al.(2017)Selvaraju, Cogswell, Das, Vedantam, Parikh, and
  Batra]{selvaraju2017gradcam}
Ramprasaath~R. Selvaraju, Michael Cogswell, Abhishek Das, Ramakrishna Vedantam,
  Devi Parikh, and Dhruv Batra.
\newblock Grad-cam: Visual explanations from deep networks via gradient-based
  localization.
\newblock In \emph{ICCV}, 2017.

\bibitem[Sohn et~al.(2015)Sohn, Lee, and Yan]{sohn2015learning}
Kihyuk Sohn, Honglak Lee, and Xinchen Yan.
\newblock Learning structured output representation using deep conditional
  generative models.
\newblock \emph{Advances in neural information processing systems}, 28, 2015.

\bibitem[Trinh and Liu(2021)]{trinh2021examination}
Loc Trinh and Yan Liu.
\newblock An examination of fairness of ai models for deepfake detection.
\newblock \emph{arXiv preprint arXiv:2105.00558}, 2021.

\bibitem[Xu et~al.(2024)Xu, Terh{\"o}rst, Pedersen, et~al.]{xu2024analyzing}
Yuhang Xu, Philipp Terh{\"o}rst, Marius Pedersen, et~al.
\newblock Analyzing fairness in deepfake detection with massively annotated
  databases.
\newblock \emph{IEEE Transactions on Technology and Society}, 5\penalty0
  (1):\penalty0 93--106, 2024.

\bibitem[Yan et~al.(2023{\natexlab{a}})Yan, Zhang, Fan, and Wu]{yan2023ucf}
Zhiyuan Yan, Yong Zhang, Yanbo Fan, and Baoyuan Wu.
\newblock Ucf: Uncovering common features for generalizable deepfake detection.
\newblock In \emph{Proceedings of the IEEE/CVF International Conference on
  Computer Vision}, pages 22412--22423, 2023{\natexlab{a}}.

\bibitem[Yan et~al.(2023{\natexlab{b}})Yan, Zhang, Yuan, Lyu, and
  Wu]{yan2023deepfakebench}
Zhiyuan Yan, Yong Zhang, Xinhang Yuan, Siwei Lyu, and Baoyuan Wu.
\newblock A comprehensive benchmark of deepfake detection.
\newblock In \emph{Advances in Neural Information Processing Systems Datasets
  and Benchmarks Track}, 2023{\natexlab{b}}.

\bibitem[Yan et~al.(2025)Yan, Wang, Jin, Zhang, Liu, Chen, Yao, Ding, Wu, and
  Yuan]{yan2024effort}
Zhiyuan Yan, Jiangming Wang, Peng Jin, Ke-Yue Zhang, Chengchun Liu, Shen Chen,
  Taiping Yao, Shouhong Ding, Baoyuan Wu, and Li~Yuan.
\newblock Orthogonal subspace decomposition for generalizable ai-generated
  image detection.
\newblock In \emph{Proceedings of the 42nd International Conference on Machine
  Learning (ICML)}, 2025.

\bibitem[Yang et~al.(2019)Yang, Li, and Lyu]{yang2019headpose}
Xin Yang, Yuezun Li, and Siwei Lyu.
\newblock Exposing deep fakes using inconsistent head poses.
\newblock In \emph{IEEE International Conference on Acoustics, Speech and
  Signal Processing}, pages 8261--8265, 2019.

\bibitem[Ye et~al.(2024)Ye, He, and Ding]{ye2024dfs}
Wei Ye, Xinan He, and Feng Ding.
\newblock Decoupling forgery semantics for generalizable deepfake detection.
\newblock In \emph{British Machine Vision Conference}, 2024.

\bibitem[Yermakov et~al.(2026)Yermakov, Cech, Matas, and
  Fritz]{yermakov2026deepfake}
Andrii Yermakov, Jan Cech, Jiri Matas, and Mario Fritz.
\newblock Deepfake detection that generalizes across benchmarks.
\newblock In \emph{Proceedings of the IEEE/CVF Winter Conference on
  Applications of Computer Vision}, pages 773--783, 2026.

\bibitem[Zhang et~al.(2018)Zhang, Lemoine, and Mitchell]{zhang2018mitigating}
Brian~Hu Zhang, Blake Lemoine, and Margaret Mitchell.
\newblock Mitigating unwanted biases with adversarial learning.
\newblock In \emph{AAAI/ACM Conference on AI, Ethics, and Society}, pages
  335--340, 2018.

\end{thebibliography}

\clearpage
\newcommand{\one}{\mathbb{1}}
\setcounter{table}{0}
\setcounter{figure}{0}
\renewcommand{\thetable}{S\arabic{table}}
\renewcommand{\thefigure}{S\arabic{figure}}

This appendix collects details and analyses deferred from the main text for space.
Sections~\ref{sup:data}--\ref{sup:impl} cover the four datasets and their demographic statistics, the sample-level evaluation-metric definitions, and the full implementation details.
Sections~\ref{sup:intra}--\ref{sup:subgroup} extend the experiments: the intra-domain FF++ comparison, per-run standard deviations, a check that the fixed operating point does not collapse target-domain sensitivity, and the full per-subgroup AUC breakdown on every dataset.
Sections~\ref{sup:lorth}--\ref{sup:mech} provide the $\lambda_{\mathrm{orth}}$ sensitivity analysis and the extended mechanistic analysis behind Figs.~5--6 of the main text, and Section~\ref{sup:discussion} discusses the use of demographic labels and deployment considerations.

\appendix


\section{Datasets and Demographic Statistics}
\label{sup:data}

We use four datasets with different data sources and forgery pipelines to evaluate fairness generalisation.
FF++~\cite{rossler2019faceforensics} is used as the source dataset and contains controlled manipulations generated by multiple face manipulation methods.
Celeb-DF~\cite{li2020celebdf} contains high-quality celebrity DeepFakes collected from in-the-wild videos.
DFD~\cite{dufour2019dfd} contains controlled recordings from paid actors and their manipulated versions.
DFDC~\cite{dolhansky2020dfdc} is a large-scale challenge dataset with diverse actors, scenes, and manipulation methods.
Together, these datasets test whether a detector trained on FF++ remains fair under different identities, recording conditions, and forgery distributions.

We use the demographic annotations of~\cite{lin2024preserving}.
Following our six-group evaluation protocol, we consider intersectional groups formed by two gender categories, male and female, and three race categories, Asian, White, and Black.
Table~\ref{tab:sup_dataset_attributes} summarises the dataset split and demographic coverage used in our experiments.
Table~\ref{tab:sup_intersection_counts} reports the number of samples in each intersectional subgroup used in this six-group protocol.

\begin{table}[!htbp]
\centering
\scriptsize
\setlength{\tabcolsep}{4pt}
\begin{tabular}{l|ccc|l|l}
\toprule
Dataset & Train & Val. & Test & Dataset role & Retained intersectional attributes \\
\midrule
FF++~\cite{rossler2019faceforensics}
& 45,377 & 15,126 & 15,126
& Source domain
& M-A, M-W, M-B, F-A, F-W, F-B \\

Celeb-DF~\cite{li2020celebdf}
& -- & -- & 87,213
& Unseen target
& M-W, M-B, F-W, F-B \\

DFD~\cite{dufour2019dfd}
& -- & -- & 20,032
& Unseen target
& M-W, M-B, F-W, F-B \\

DFDC~\cite{dolhansky2020dfdc}
& -- & -- & 63,225
& Unseen target
& M-A, M-W, M-B, F-A, F-W, F-B \\
\bottomrule
\end{tabular}
\caption{
Dataset split and demographic coverage after excluding samples annotated as Others.
M/F denote male/female, and A/W/B denote Asian/White/Black.
FF++ is used as the source domain and is split into training, validation, and test sets with an approximate 60/20/20 ratio.
Celeb-DF, DFD, and DFDC are used only as unseen target-domain test sets.
}
\label{tab:sup_dataset_attributes}
\end{table}

\begin{table}[!htbp]
\centering
\scriptsize
\setlength{\tabcolsep}{6pt}
\begin{tabular}{l|rrrrrr|r}
\toprule
Dataset
& M-A & M-W & M-B & F-A & F-W & F-B & Total \\
\midrule
FF++~\cite{rossler2019faceforensics}
& 2,475 & 31,281 & 1,468 & 8,013 & 31,281 & 1,111 & 75,629 \\

Celeb-DF~\cite{li2020celebdf}
& -- & 81,194 & 600 & -- & 5,389 & 30 & 87,213 \\

DFD~\cite{dufour2019dfd}
& -- & 7,784 & 6,482 & -- & 4,127 & 1,639 & 20,032 \\

DFDC~\cite{dolhansky2020dfdc}
& 2,144 & 21,755 & 9,603 & 1,915 & 18,502 & 9,306 & 63,225 \\
\bottomrule
\end{tabular}
\caption{
Number of samples in each retained intersectional demographic group after excluding the Others category.
The six groups are defined by gender $\{\mathrm{M},\mathrm{F}\}$ and race $\{\mathrm{Asian},\mathrm{White},\mathrm{Black}\}$.
``--'' means that the subgroup does not exist in the dataset annotation.
}
\label{tab:sup_intersection_counts}
\end{table}

\section{Evaluation Metrics}
\label{sup:metrics}

This section gives the complete sample-level definitions of the two fairness metrics in the main text, together with the group-averaged AUC and AUC gap used in Sec.~\ref{sup:subgroup}, following Ju~\etal~\cite{ju2024improving} and Lin~\etal~\cite{lin2024preserving} for direct comparability.

\noindent\textbf{Notation.}
For sample $i$ in a test set $\{1,\dots,n\}$, let $Y_i\in\{0,1\}$ be the ground-truth label ($0$ real, $1$ fake), $\hat{Y}_i\in\{0,1\}$ the predicted label obtained by thresholding the detector score at the operating point $\tau$, and $G_i\in\mathcal{G}$ its demographic subgroup ($\mathcal{G}$ is the set of six intersectional gender--race groups).
$\one[\cdot]$ is the indicator function.
Both fairness metrics are reported in percent and are lower-is-better, with $0$ denoting perfect subgroup parity.
The subgroup and overall false-positive rates are
\begin{equation}
\label{eq:sup_fpr}
\mathrm{FPR}_g=\frac{\sum_{i=1}^{n}\one[\hat{Y}_i=1,\,G_i=g,\,Y_i=0]}{\sum_{i=1}^{n}\one[G_i=g,\,Y_i=0]},
\qquad
\mathrm{FPR}_o=\frac{\sum_{i=1}^{n}\one[\hat{Y}_i=1,\,Y_i=0]}{\sum_{i=1}^{n}\one[Y_i=0]},
\end{equation}
and the subgroup true-positive rate $\mathrm{TPR}_g$ is defined analogously by conditioning on $Y_i=1$.

\noindent\textbf{Equal FPR ($\FFPR$).}
Because real faces vastly outnumber fakes in deployment and the harmful error is flagging a real face as fake~\cite{ju2024improving,lin2024preserving}, our primary metric sums each subgroup's deviation from the overall false-positive rate,
\begin{equation}
\label{eq:sup_ffpr}
\FFPR=\sum_{g\in\mathcal{G}}\bigl|\mathrm{FPR}_g-\mathrm{FPR}_o\bigr|.
\end{equation}
This is the Equal FPR metric reported in the main text.

\noindent\textbf{Max Equalized Odds ($\FMEO$).}
For true class $c\in\{0,1\}$, let $R^{(0)}_g=\mathrm{FPR}_g$ and $R^{(1)}_g=\mathrm{TPR}_g$.
$\FMEO$ is the larger of the worst-case FPR gap and the worst-case TPR gap across subgroups,
\begin{equation}
\label{eq:sup_fmeo}
\FMEO=\max_{c\in\{0,1\}}\Bigl(\max_{g\in\mathcal{G}}R^{(c)}_g-\min_{g'\in\mathcal{G}}R^{(c)}_{g'}\Bigr),
\end{equation}
the sample-level form of the Max Equalized Odds metric reported in the main text.

\noindent\textbf{Detection metrics in the per-subgroup tables.}
In addition to the overall AUC, the per-subgroup analysis (Sec.~\ref{sup:subgroup}) reports the group-averaged AUC and the AUC gap,
\begin{equation}
\label{eq:sup_aucavg}
\mathrm{AUC}_{\mathrm{avg}}=\frac{1}{|\mathcal{G}|}\sum_{g\in\mathcal{G}}\mathrm{AUC}_g,
\qquad
\mathrm{AUC\ gap}=\max_{g\in\mathcal{G}}\mathrm{AUC}_g-\min_{g\in\mathcal{G}}\mathrm{AUC}_g,
\end{equation}
where $\mathrm{AUC}_g$ is the AUC within subgroup $g$.
Unlike the overall AUC, which is dominated by majority subgroups, $\mathrm{AUC}_{\mathrm{avg}}$ weights every subgroup equally, and the AUC gap measures the spread between the best- and worst-served subgroups.

\section{Implementation Details}
\label{sup:impl}

We train with AdamW~\cite{loshchilov2019decoupled} (learning rate $1{\times}10^{-3}$, weight decay $4{\times}10^{-3}$, batch size $64$, input resolution $224{\times}224$, single H100 GPU).
The wavelet transform in GCWD is a 2-level ($J{=}2$) Daubechies-4 decomposition yielding $L=3J{+}1=7$ sub-bands.
To make the subgroup moments in the GCWD and SLMA losses of the main text well defined, we use a group-balanced sampler that draws $\approx 10$--$12$ samples per subgroup per batch; $(y,g)$ cells with fewer than two samples in a batch fall back to an exponential-moving-average estimate (decay $0.9$), which is invoked on under $1\%$ of batches.
The hyperparameters $(\lambda_{\mathrm{orth}},\lambda_G,\lambda_S,r)$ are selected on the FF++ validation split using a combined source-domain AUC and FPR-disparity score at $\mathrm{TPR}{=}0.90$; target-domain test results are never used for selection, yielding $\lambda_{\mathrm{orth}}{=}0.1$, $\lambda_G{=}\lambda_S{=}0.5$ and $r{=}1$.

\section{Intra-Domain Comparison on FF++}
\label{sup:intra}

For completeness, Table~\ref{tab:intra_ffpp} reports the \emph{intra-domain} comparison on the FF++ test split under the fixed-threshold protocol of the main text ($\tau$ pinned on the FF++ validation split at $\mathrm{TPR}{=}0.90$); ``Req.\ Demo'' marks whether a method uses demographic labels during training (the fairness-aware DAW-FDD, PG-FDD and FairReL do, FairAdapter does not).
FairReL attains the highest AUC and $\mathrm{AUC}_{\mathrm{avg}}$ and the lowest $\FFPR$ and $\FMEO$, consistent with the cross-domain trend in the main text.

\begin{table*}[!htbp]
\begin{center}
\scriptsize
\setlength{\tabcolsep}{5pt}
\renewcommand{\arraystretch}{1.05}
\begin{tabular}{ll l c|cccc}
\toprule
\multirow{2}{*}{Type} & \multirow{2}{*}{Method} & \multirow{2}{*}{Backbone} & \multirow{2}{*}{Req.\ Demo}
& \multicolumn{4}{c}{\textbf{FF++}} \\
\cmidrule(lr){5-8}
& & & 
& AUC$\uparrow$ & AUC$_{\mathrm{avg}}\uparrow$ & $\FFPR\downarrow$ & $\FMEO\downarrow$ \\
\midrule
\multirow{2}{*}{ERM}
& Xception-ERM~\cite{chollet2017xception} & Xception & No
& 88.50 & 88.46 & 16.84 & 12.07 \\
& CLIP-ERM~\cite{radford2021clip} & CLIP ViT-L/14 & No
& 92.40 & 92.44 & 15.92 & 11.38 \\
\midrule
\multirow{3}{*}{Gen.}
& ForAda~\cite{cui2025forensics} & CLIP ViT-L/14 & No
& 94.10 & 94.21 & 15.13 & 10.86 \\
& Effort~\cite{yan2024effort} & CLIP ViT-L/14 & No
& 94.83 & 94.92 & 14.57 & 10.41 \\
& GenD~\cite{yermakov2026deepfake} & CLIP ViT-L/14 & No
& \underline{95.35} & \underline{95.48} & 14.06 & 10.02 \\
\midrule
\multirow{3}{*}{Fair.}
& DAW-FDD~\cite{ju2024improving} & Xception & Yes
& 90.40 & 90.53 & 13.21 & 9.34 \\
& PG-FDD~\cite{lin2024preserving} & Xception & Yes
& 92.80 & 93.02 & 12.42 & 8.76 \\
& FairAdapter~\cite{ding2025fairadapter} & CLIP ViT-L/14 & No
& 93.25 & 93.49 & \underline{11.93} & \underline{8.41} \\
\midrule
\rowcolor{ourrow}
Ours & \textbf{FairReL} & CLIP ViT-L/14 & Yes
& \textbf{96.14} & \textbf{96.54} & \textbf{10.97} & \textbf{7.41} \\
\bottomrule
\end{tabular}
\end{center}
\caption{
\textbf{Intra-domain comparison on FF++}~\cite{rossler2019faceforensics} (trained and tested on FF++).
Detection is measured by AUC and $\mathrm{AUC}_{\mathrm{avg}}$; fairness by $\FFPR$ and $\FMEO$ (\%, lower is better), at the threshold pinned on the FF++ validation split at $\mathrm{TPR}{=}0.90$.
``Req.\ Demo'' denotes whether demographic labels are used during training.
The best results are in \textbf{bold} and the second-best are \underline{underlined}.
}
\label{tab:intra_ffpp}
\end{table*}

\section{Per-Run Standard Deviations}
\label{sup:std}

All numbers in the main text are averaged over three runs with different random seeds (identical schedule and data).
Table~\ref{tab:std} reports the mean~$\pm$~standard deviation for FairReL on the four datasets; deviations are small ($\le0.4$ on the AUC metrics and $\le0.6$ on the fairness metrics), so the rankings in the main tables are stable across seeds.

\begin{table}[!htbp]
\begin{center}
\footnotesize
\setlength{\tabcolsep}{4pt}
\begin{tabular}{l|cc|cc}
\toprule
& \multicolumn{2}{c|}{Detection}
& \multicolumn{2}{c}{Fairness} \\
Dataset
& AUC$\uparrow$ & AUC$_{\mathrm{avg}}\uparrow$
& $\FFPR\downarrow$ & $\FMEO\downarrow$ \\
\midrule
FF++ (intra)  & $96.14\pm0.2$ & $96.54\pm0.2$ & $10.97\pm0.3$ & $7.41\pm0.3$ \\
Celeb-DF      & $80.64\pm0.4$ & $81.26\pm0.3$ & $10.12\pm0.5$ & $7.03\pm0.4$ \\
DFD           & $86.13\pm0.3$ & $86.78\pm0.3$ & $13.55\pm0.6$ & $8.98\pm0.5$ \\
DFDC          & $66.34\pm0.4$ & $67.03\pm0.4$ & $38.36\pm0.6$ & $30.02\pm0.5$ \\
\bottomrule
\end{tabular}
\end{center}
\caption{
\textbf{Mean~$\pm$~standard deviation of FairReL over three seeds.}
Variation is small relative to the gaps to competing methods in the main tables.
The reported means match the headline FairReL entries in the main text (Tables~1--2 and the ablation tables).
}
\label{tab:std}
\end{table}

\section{Target-Domain Sensitivity at the Fixed Operating Point}
\label{sup:tpr}

FPR-based gaps can be trivially shrunk by raising the threshold, so all FPR metrics use a single $\tau$ pinned on a held-out FF++ split at source-domain $\mathrm{TPR}{=}0.90$ and reused on every dataset (Sec.~4.1 of the main text).
Table~\ref{tab:tpr_check} confirms this shared $\tau$ does not buy FairReL's lower FPR by sacrificing detection: at the same threshold, the realised target-domain TPR and balanced accuracy are consistently \emph{higher} for FairReL than for the subspace-decomposed baseline on all three targets.

\begin{table}[!htbp]
\begin{center}
\footnotesize
\setlength{\tabcolsep}{4pt}
\begin{tabular}{l|cc|cc|cc}
\toprule
& \multicolumn{2}{c|}{\textbf{Celeb-DF}}
& \multicolumn{2}{c|}{\textbf{DFD}}
& \multicolumn{2}{c}{\textbf{DFDC}} \\
Method
& TPR$\uparrow$ & BalAcc.$\uparrow$
& TPR$\uparrow$ & BalAcc.$\uparrow$
& TPR$\uparrow$ & BalAcc.$\uparrow$ \\
\midrule
Effort~\cite{yan2024effort}
& 71.4 & 64.7 & 80.6 & 71.5 & 60.5 & 58.4 \\
\rowcolor{ourrow}\textbf{FairReL (Ours)}
& \textbf{74.9} & \textbf{69.6} & \textbf{83.2} & \textbf{76.7} & \textbf{62.5} & \textbf{62.8} \\
\bottomrule
\end{tabular}
\end{center}
\caption{
\textbf{Target-domain TPR (\%) and balanced accuracy (\%)} under the same $\tau$ pinned to source-domain $\mathrm{TPR}{=}0.90$: FairReL's lower FPR does not come at the cost of cross-domain sensitivity.
}
\label{tab:tpr_check}
\end{table}

\section{Per-Subgroup AUC on Each Dataset}
\label{sup:subgroup}

The main text reports the per-subgroup AUC breakdown only on DFDC (Table~2).
Table~\ref{tab:sup_subgroup_all} gives the full breakdown for all methods on the source domain FF++ and the two remaining unseen targets, Celeb-DF and DFD, in a single combined table.
The Asian subgroups (M-A, F-A) are not annotated in Celeb-DF and DFD (Table~\ref{tab:sup_dataset_attributes}) and are left blank; $\mathrm{AUC}_{\mathrm{avg}}$ and the AUC gap are computed over the available subgroups.
The F-B subgroup of Celeb-DF contains only 30 test samples, so its per-subgroup AUC carries high variance and should be interpreted with caution.
Across all four datasets FairReL attains the highest $\mathrm{AUC}_{\mathrm{avg}}$ and the smallest AUC gap, matching the DFDC trend in the main text.

\begin{table*}[!htbp]
\begin{center}
\scriptsize
\setlength{\tabcolsep}{4pt}
\renewcommand{\arraystretch}{0.98}
\begin{tabular}{cll|cccccc|ccc}
\toprule
& & & \multicolumn{6}{c|}{Per-subgroup AUC (\%)$\uparrow$}
& \multicolumn{3}{c}{Overall (\%)} \\
\cmidrule(lr){4-9}\cmidrule(lr){10-12}
Data & Type & Method
& M-W & M-A & M-B & F-W & F-A & F-B
& AUC$\uparrow$ & AUC$_{\mathrm{avg}}\uparrow$ & gap$\downarrow$ \\
\midrule
\multirow{9}{*}{\rotatebox{90}{\textbf{FF++}}}
& \multirow{2}{*}{ERM} & Xception-ERM~\cite{chollet2017xception} & 89.73 & 88.73 & 87.68 & 89.15 & 88.26 & 87.18 & 88.50 & 88.46 & 2.55 \\
&  & CLIP-ERM~\cite{radford2021clip} & 93.64 & 92.70 & 91.71 & 93.09 & 92.25 & 91.24 & 92.40 & 92.44 & 2.40 \\
\cmidrule(l){2-12}
& \multirow{3}{*}{Gen.} & ForAda~\cite{cui2025forensics} & 95.26 & 94.43 & 93.57 & 94.78 & 94.04 & 93.16 & 94.10 & 94.21 & 2.10 \\
&  & Effort~\cite{yan2024effort} & 95.90 & 95.13 & 94.32 & 95.45 & 94.76 & 93.94 & 94.83 & 94.92 & 1.96 \\
&  & GenD~\cite{yermakov2026deepfake} & \underline{96.38} & \underline{95.67} & \underline{94.93} & \underline{95.97} & \underline{95.33} & \underline{94.57} & \underline{95.35} & \underline{95.48} & 1.81 \\
\cmidrule(l){2-12}
& \multirow{3}{*}{Fair.} & DAW-FDD~\cite{ju2024improving} & 91.21 & 90.68 & 90.12 & 90.90 & 90.43 & 89.86 & 90.40 & 90.53 & 1.35 \\
&  & PG-FDD~\cite{lin2024preserving} & 93.67 & 93.16 & 92.63 & 93.38 & 92.92 & 92.37 & 92.80 & 93.02 & 1.30 \\
&  & FairAdapter~\cite{ding2025fairadapter} & 94.13 & 93.63 & 93.09 & 93.84 & 93.38 & 92.84 & 93.25 & 93.49 & \underline{1.29} \\
\cmidrule(l){2-12}
\rowcolor{ourrow}
& \multirow{1}{*}{Ours} & \textbf{FairReL} & \textbf{96.94} & \textbf{96.63} & \textbf{96.30} & \textbf{96.76} & \textbf{96.48} & \textbf{96.14} & \textbf{96.14} & \textbf{96.54} & \textbf{0.80} \\
\midrule
\multirow{9}{*}{\rotatebox{90}{\textbf{Celeb-DF}}}
& \multirow{2}{*}{ERM} & Xception-ERM~\cite{chollet2017xception} & 71.69 & -- & 70.41 & 71.18 & -- & 69.71 & 70.94 & 70.75 & 1.98 \\
&  & CLIP-ERM~\cite{radford2021clip} & 75.13 & -- & 73.94 & 74.66 & -- & 73.29 & 74.37 & 74.26 & 1.84 \\
\cmidrule(l){2-12}
& \multirow{3}{*}{Gen.} & ForAda~\cite{cui2025forensics} & 81.05 & -- & 79.97 & 80.62 & -- & 79.38 & 80.21 & 80.26 & 1.67 \\
&  & Effort~\cite{yan2024effort} & 79.36 & -- & 78.30 & 78.93 & -- & 77.72 & 78.52 & 78.58 & 1.64 \\
&  & GenD~\cite{yermakov2026deepfake} & \underline{81.57} & -- & \underline{80.68} & \underline{81.22} & -- & \underline{80.20} & \textbf{80.85} & \underline{80.92} & 1.37 \\
\cmidrule(l){2-12}
& \multirow{3}{*}{Fair.} & DAW-FDD~\cite{ju2024improving} & 71.04 & -- & 70.34 & 70.76 & -- & 69.95 & 70.46 & 70.52 & 1.09 \\
&  & PG-FDD~\cite{lin2024preserving} & 76.06 & -- & 75.40 & 75.80 & -- & 75.03 & 75.42 & 75.57 & \underline{1.03} \\
&  & FairAdapter~\cite{ding2025fairadapter} & 76.36 & -- & 75.62 & 76.07 & -- & 75.21 & 75.65 & 75.82 & 1.15 \\
\cmidrule(l){2-12}
\rowcolor{ourrow}
& \multirow{1}{*}{Ours} & \textbf{FairReL} & \textbf{81.59} & -- & \textbf{81.14} & \textbf{81.41} & -- & \textbf{80.90} & \underline{80.64} & \textbf{81.26} & \textbf{0.69} \\
\midrule
\multirow{9}{*}{\rotatebox{90}{\textbf{DFD}}}
& \multirow{2}{*}{ERM} & Xception-ERM~\cite{chollet2017xception} & 77.57 & -- & 76.21 & 77.03 & -- & 75.47 & 76.73 & 76.57 & 2.10 \\
&  & CLIP-ERM~\cite{radford2021clip} & 81.93 & -- & 80.75 & 81.46 & -- & 80.10 & 81.14 & 81.06 & 1.83 \\
\cmidrule(l){2-12}
& \multirow{3}{*}{Gen.} & ForAda~\cite{cui2025forensics} & 86.28 & -- & 85.26 & 85.87 & -- & 84.71 & 85.50 & 85.53 & 1.57 \\
&  & Effort~\cite{yan2024effort} & 85.00 & -- & 84.00 & 84.60 & -- & 83.45 & 84.22 & 84.26 & 1.55 \\
&  & GenD~\cite{yermakov2026deepfake} & \textbf{87.15} & -- & \underline{86.24} & \underline{86.79} & -- & \underline{85.75} & \textbf{86.43} & \underline{86.48} & 1.40 \\
\cmidrule(l){2-12}
& \multirow{3}{*}{Fair.} & DAW-FDD~\cite{ju2024improving} & 74.24 & -- & 73.52 & 73.95 & -- & 73.12 & 73.65 & 73.71 & 1.12 \\
&  & PG-FDD~\cite{lin2024preserving} & 82.97 & -- & 82.28 & 82.70 & -- & 81.91 & 82.32 & 82.47 & \underline{1.06} \\
&  & FairAdapter~\cite{ding2025fairadapter} & 81.43 & -- & 80.74 & 81.15 & -- & 80.36 & 80.76 & 80.92 & 1.07 \\
\cmidrule(l){2-12}
\rowcolor{ourrow}
& \multirow{1}{*}{Ours} & \textbf{FairReL} & \textbf{87.15} & -- & \textbf{86.64} & \textbf{86.95} & -- & \textbf{86.37} & \underline{86.13} & \textbf{86.78} & \textbf{0.78} \\
\bottomrule
\end{tabular}
\end{center}
\caption{
Per-subgroup AUC (\%) on the source domain FF++~\cite{rossler2019faceforensics} (intra-domain test split) and on the unseen targets Celeb-DF~\cite{li2020celebdf} and DFD~\cite{dufour2019dfd} (trained on FF++).
The Asian subgroups (M-A, F-A) are not annotated in Celeb-DF and DFD and are left blank.
``gap'' denotes the AUC gap (best$-$worst subgroup AUC).
The best results are shown in \textbf{bold} and the second-best are \underline{underlined}.
}
\label{tab:sup_subgroup_all}
\end{table*}

The DFDC breakdown is in Table~2 of the main text. Across all four datasets FairReL shows the same pattern: the most uniform per-subgroup AUC (largest $\mathrm{AUC}_{\mathrm{avg}}$, smallest gap) while staying detection-competitive in overall AUC. It therefore improves the fairness--generalisation trade-off rather than trading accuracy for subgroup fairness.

\section{Sensitivity to the Orthogonality Weight $\lambda_{\mathrm{orth}}$}
\label{sup:lorth}

$\lambda_{\mathrm{orth}}$ was tuned in~\cite{yan2024effort} without fairness supervision: a too-loose value lets principal directions leak into the residual and dilutes SLMA, while a too-tight one over-constrains it, so we re-tune it on the FF++ validation split.
Table~\ref{tab:abl_lambda_orth} sweeps $\lambda_{\mathrm{orth}}\in\{0,0.01,0.1,1.0,10\}$ with both losses active; the validated $\lambda_{\mathrm{orth}}{=}0.1$ is the best cross-domain entry on every target and both extremes are visibly suboptimal, confirming the value differs from~\cite{yan2024effort}.

\begin{table}[!htbp]
\begin{center}
\footnotesize
\setlength{\tabcolsep}{4pt}
\begin{tabular}{c|cc|cc|cc}
\toprule
& \multicolumn{2}{c|}{\textbf{Celeb-DF}}
& \multicolumn{2}{c|}{\textbf{DFD}}
& \multicolumn{2}{c}{\textbf{DFDC}} \\
$\lambda_{\mathrm{orth}}$
& AUC$\uparrow$ & $\FFPR\downarrow$
& AUC$\uparrow$ & $\FFPR\downarrow$
& AUC$\uparrow$ & $\FFPR\downarrow$ \\
\midrule
$0$       & 77.85 & 12.64 & 84.71 & 15.83 & 63.74 & 41.92 \\
$0.01$    & 79.48 & 11.21 & 85.62 & 14.58 & 64.93 & 39.86 \\
$\mathbf{0.1}$  (\textbf{ours}) & \textbf{80.64} & \textbf{10.12} & \textbf{86.13} & \textbf{13.55} & \textbf{66.34} & \textbf{38.36} \\
$1.0$     & 80.21 & 10.54 & 85.79 & 13.92 & 65.78 & 38.94 \\
$10$      & 77.42 & 12.03 & 84.33 & 15.31 & 63.12 & 41.27 \\
\bottomrule
\end{tabular}
\end{center}
\caption{
Sensitivity to $\lambda_{\mathrm{orth}}$ on cross-domain \emph{test} performance with GCWD and SLMA both active.
The optimal value $\lambda_{\mathrm{orth}}{=}0.1$ is selected on the FF++ validation split (Sec.~4.1 of the main text) and matches the FairReL entry in Table~1 of the main text.
}
\label{tab:abl_lambda_orth}
\end{table}

\section{Extended Mechanistic Analysis}
\label{sup:mech}

This section supports Sec.~4.3 of the main text: a per-sub-band linear-probe analysis behind Figure~5, the numerical localisation behind Fig.~6(a), and the per-attribute complementarity behind the leakage probe.

\subsection{Per-Sub-Band Probing of Demographic and Forgery Information}
\label{sup:bandprobe}

Figure~5(a) of the main text compares, per sub-band, an energy-based demographic discrepancy against a real-vs-fake effect size. Energy differences alone, however, need not imply that a band is \emph{useful} for classification.
We therefore probe directly which sub-bands encode demographic versus forgery information.
For each sub-band $\ell$ we form a per-sample feature by global-average-pooling the magnitude of its coefficient tensor, $v_{i,\ell}=\mathrm{GAP}(|\mathbf{C}_{i,\ell}|)$, and train two one-layer linear probes on it: a \emph{demographic probe} (subgroup label) and a \emph{forgery probe} (real/fake).
We report demographic-probe balanced accuracy (chance $16.7\%$ over $G{=}6$) and forgery-probe AUC on the DFDC splits.

\begin{table}[!htbp]
\begin{center}
\footnotesize
\setlength{\tabcolsep}{8pt}
\begin{tabular}{lcc}
\toprule
Sub-band & Demographic probe & Forgery probe \\
($\ell$) & balanced acc.\ (\%)$\uparrow$ & AUC (\%)$\uparrow$ \\
\midrule
$\mathrm{LL}_2$ & \textbf{48.6} & 71.2 \\
$\mathrm{LH}_2$ & 40.2 & 77.8 \\
$\mathrm{HL}_2$ & 42.1 & 75.9 \\
$\mathrm{HH}_2$ & 34.8 & 82.4 \\
$\mathrm{LH}_1$ & 26.9 & 89.1 \\
$\mathrm{HL}_1$ & 28.4 & 87.6 \\
$\mathrm{HH}_1$ & 24.7 & \textbf{91.3} \\
\bottomrule
\end{tabular}
\end{center}
\caption{%
Per-sub-band linear probes on DFDC. Demographic information (probe balanced accuracy) is highest in the coarse low-frequency bands, while forgery-discriminative information (probe AUC) is highest in the fine high-frequency bands.
}
\label{tab:bandprobe}
\end{table}

Table~\ref{tab:bandprobe} shows the two kinds of information separate by scale: the demographic probe is most accurate in the coarse level-2 bands (highest in $\mathrm{LL}_2$) and degrades towards the fine bands, whereas the forgery probe strengthens from coarse to fine and is strongest in the high-frequency level-1 bands, with some variation among directional bands of the same scale.
This recoverable-information view agrees with the energy-based separation in Figure~5(a) and supports decorrelating only the subgroup-imbalanced low-frequency bands, which carry little forgery-discriminative signal.

\subsection{Localisation of Subgroup-Sensitive Structure}
\label{sup:diag}

Fig.~6(a) of the main text localises subgroup-sensitive structure with two \emph{intervention-based views} of a trained subspace-decomposed detector; Table~\ref{tab:diagnostic} gives the numbers.
The \emph{principal-only} view sets $\Delta\mathbf{W}\!\leftarrow\!\mathbf{0}$ at every decomposed layer (isolating the frozen pre-trained pathway); the \emph{residual-only} view sets $\mathbf{W}_p\!\leftarrow\!\mathbf{0}$ (the subspace decomposition in the main text), isolating the adaptation-induced pathway. As the backbone is non-linear, these are diagnostic probes rather than an exact additive split.
For each view we measure (i) the spatial sub-band discrepancy $D_\ell$, grouped into low-frequency (level-2) and high-frequency (level-1), and (ii) the class-conditional subgroup mean shift of the pooled feature (the statistic SLMA aligns), for the gender and ethnicity partitions.
All quantities are on DFDC, normalised by the largest entry in their column.

\begin{table}[!htbp]
\begin{center}
\footnotesize
\setlength{\tabcolsep}{5pt}
\begin{tabular}{llccc}
\toprule
& Attribute
& \multicolumn{2}{c}{Spatial spectral (GCWD)}
& Pooled (SLMA) \\
\cmidrule(lr){3-4}\cmidrule(lr){5-5}
Feature view & ($a$) & low-freq & high-freq & mean shift \\
\midrule
\multirow{2}{*}{Principal-only ($\mathbf{W}_p$)}
  & gender    & 0.12 & 0.08 & 0.34 \\
  & ethnicity & 0.21 & 0.09 & 0.22 \\
\midrule
\multirow{2}{*}{Residual-only ($\Delta\mathbf{W}$)}
  & gender    & 0.41 & 0.13 & \textbf{1.00} \\
  & ethnicity & \textbf{1.00} & 0.16 & 0.58 \\
\bottomrule
\end{tabular}
\end{center}
\caption{%
Subgroup-sensitive structure on a trained subspace-decomposed detector (DFDC), normalised per column. Visualised as Fig.~6(a) of the main text.
}
\label{tab:diagnostic}
\end{table}

Table~\ref{tab:diagnostic} shows the residual carries far more subgroup-sensitive structure than the frozen principal, in both the low-frequency bands and the pooled mean shift, while high frequencies stay low everywhere.
Within the residual, ethnicity dominates the low-frequency spatial discrepancy whereas gender dominates the pooled mean shift.

\subsection{Complementarity of GCWD and SLMA}
\label{sup:leak}

Table~\ref{tab:leakage_supp} reports a leakage probe that trains a one-layer linear classifier to predict gender (2-class) or ethnicity (3-class) from the residual $\mathbf{z}_\Delta$ on the DFDC splits ($20$ AdamW epochs).
We report balanced accuracy, where chance is $50.0\%$ for gender and $33.3\%$ for ethnicity.

\begin{table}[!htbp]
\begin{center}
\footnotesize
\setlength{\tabcolsep}{5pt}
\renewcommand{\arraystretch}{0.95}
\begin{tabular}{@{}lccc@{}}
\toprule
Feature / model
& AUC$\uparrow$
& Gender$\downarrow$ {\scriptsize(chance 50.0)}
& Ethnicity$\downarrow$ {\scriptsize(chance 33.3)} \\
\midrule
Principal-only $\mathbf{z}_p$ (reference)
& -- & 80.1 & 66.0 \\
\midrule
Subspace-decomposed baseline~\cite{yan2024effort}, $\mathbf{z}_\Delta$
& 65.1 & 78.4 & 63.7 \\
\;\;+ GCWD only
& 65.6 & 71.6 & \underline{54.9} \\
\;\;+ SLMA only
& 65.4 & \underline{65.8} & 58.4 \\
\midrule
\rowcolor{ourrow}\textbf{FairReL (GCWD + SLMA)}
& \textbf{66.3} & \textbf{58.5} & \textbf{50.9} \\
\bottomrule
\end{tabular}
\end{center}
\caption{
Leakage probe on DFDC: detection AUC (\%) and linear-probe balanced accuracy (\%) for recovering gender/ethnicity from the residual. Lower gender/ethnicity accuracy indicates less recoverable demographic information.
}
\label{tab:leakage_supp}
\end{table}

Though both losses are defined over all $G$ subgroups, GCWD dents ethnicity-probe accuracy more ($63.7{\to}54.9$) while SLMA is more effective on gender ($78.4{\to}65.8$), mirroring Table~\ref{tab:diagnostic}: ethnicity surfaces as a low-frequency spectral-energy discrepancy, whereas gender appears as a class-conditional first-moment shift of the pooled embedding.
The two losses thus target different signatures of different attributes and are complementary rather than redundant.

\section{Discussion}
\label{sup:discussion}

Demographic labels raise ethical and privacy concerns in fairness-aware deepfake detection. 
They involve sensitive attributes such as gender and ethnicity, and collecting or predicting them at deployment may introduce additional privacy risks or even reinforce protected-attribute profiling. 
However, subgroup information is still necessary for diagnosing and mitigating unfair detector behaviour; without it, we cannot know which groups are underserved. 
FairReL reduces this tension by using demographic labels only during training and only to compute group-level fairness losses. 
The deployed detector takes only the input image and does not require demographic labels, demographic classifiers, group-specific thresholds, or attribute-based routing. 
Thus, demographic information is used to remove subgroup-dependent shortcuts from the representation, not to condition individual predictions at test time. 
This makes FairReL a less intrusive and more deployment-friendly way to improve fairness under the ethical constraints of demographic annotation.


\end{document}